\documentclass[11pt, a4paper, logo, copyright, nonumbering]{map}
\usepackage{CJKutf8}
\usepackage{xargs}  
\usepackage{todonotes}
\usepackage{amsmath}
\usepackage{dsfont}

\usepackage{longtable}

\usepackage{svg}
\usepackage{fontawesome}
\usepackage{array}
\usepackage{tabularx}
\usepackage{latexsym}
\usepackage{graphicx}
\usepackage[utf8]{inputenc} 
\usepackage[utf8]{inputenc} 
\usepackage{amssymb} 
\usepackage{url}            
\usepackage{amsfonts}       
\usepackage{nicefrac}       
\usepackage{microtype}      
\usepackage{xspace}

\usepackage{listings}
\usepackage{makecell}
\usepackage{inconsolata}
\usepackage{multicol}
\usepackage{amstext}

\definecolor{darkblue}{RGB}{84, 112, 198}

\definecolor{lightblue}{rgb}{0.85, 0.95, 1.0}    
\definecolor{lightgreen}{rgb}{0.90, 1.0, 0.90}    
\definecolor{lightorange}{rgb}{1.0, 0.95, 0.85}   
\definecolor{lightpurple}{rgb}{0.95, 0.90, 1.0}   
\definecolor{lightgray}{rgb}{0.97, 0.97, 0.97}    

\usepackage{tikz}
\usetikzlibrary{calc}
\definecolor{battery-empty}{rgb}{0.9, 0.9, 0.9}
\newcommand{\difficultybar}[1]{%
  \begin{tikzpicture}[baseline, scale=0.5, every node/.style={scale=0.8}]
    \foreach \i in {1,2,3,4,5} {
      \ifnum\i>#1
        \draw[fill=battery-empty] (\i*0.5-0.5, 0) rectangle (\i*0.5, 0.25);
      \else
        \pgfmathsetmacro{\colorlevel}{80 - 12*(\i)} 
        \edef\x{\noexpand\draw[fill=blue!\colorlevel!white, opacity=0.9] (\i*0.5-0.5, 0) rectangle (\i*0.5, 0.25);}
        \x
        \draw[blue!50!black] (\i*0.5-0.5, 0) rectangle (\i*0.5, 0.25);
      \fi
    }
    \fill[battery-empty!70] (2.5, 0.08) rectangle (2.6, 0.17);
    \draw[battery-empty!70!black] (2.5, 0.08) rectangle (2.6, 0.17);
  \end{tikzpicture}%
}
\renewcommand{\arraystretch}{0.96}

\definecolor{hidden-draw}{RGB}{20,68,106}
\definecolor{hidden-pink}{RGB}{255,245,247}
\usepackage[edges]{forest}

\usepackage{bm}

\usepackage{natbib}
\usepackage{CJKutf8}
\usepackage{xargs}
\usepackage{todonotes}
\usepackage{multirow}
\usepackage{cleveref}
\usepackage{subcaption}
\usepackage{url}
\usepackage{svg}
\usepackage{fontawesome}
\usepackage{xcolor}
\usepackage{float}

\usepackage[utf8]{inputenc}
\usepackage{booktabs}
\usepackage{graphicx}
\usepackage{array}
\usepackage{tabularx}
\usepackage{xcolor,colortbl}  
\definecolor{boxcolor}{HTML}{d92523} 
\definecolor{bulbcolor}{HTML}{e3b87f} 
\usepackage{url}
\usepackage{ragged2e}   
\usepackage{makecell} 

\usepackage{hyperref}       
\usepackage{url}            
\usepackage{booktabs}       
\usepackage{nicefrac}       
\usepackage{microtype}      
\usepackage{xcolor}         
\usepackage{graphicx}
\usepackage{longtable}
\usepackage{array}
\usepackage{pbox}
\usepackage{tabularx}
\usepackage{multirow}
\usepackage{wrapfig}
\usepackage{pifont}
\usepackage{lipsum}
\usepackage{subcaption}
\usepackage{enumitem}
\usepackage{tikz}
\usepackage{xstring}

\usepackage{color-edits}

\usepackage{booktabs}
\usepackage{multirow}
\usepackage[table]{xcolor}
\usepackage{tabularx}

\usepackage{parskip} 

\usepackage{threeparttable} 

\usepackage{algorithm}
\usepackage{algpseudocode}

\usepackage{xspace}
\usepackage{times}
\usepackage{latexsym}
\usepackage{booktabs}
\usepackage{fvextra}
\usepackage{graphicx}
\usepackage{subcaption} 
\usepackage[T1]{fontenc}
\usepackage{xcolor}
\usepackage[utf8]{inputenc}

\usepackage{microtype}

\usepackage{inconsolata}
\usepackage{amsmath}
\usepackage{booktabs}
\usepackage{graphicx}
\usepackage{pifont}
\usepackage{multirow}          
\usepackage{colortbl}          
\usepackage{algorithm}
\usepackage{algpseudocode}
\usepackage{xcolor}
\usepackage{amssymb}

\definecolor{rliableolive}{HTML}{BBCC33}
\definecolor{rliableblue}{HTML}{77AADD}
\definecolor{rliablered}{HTML}{f63c44}

\algrenewcommand\algorithmicrequire{\textbf{Input:}}
\algrenewcommand\algorithmicensure{\textbf{Output:}}
\algrenewcommand\alglinenumber[1]{\small #1:}

\newcommand{\ie}{\textit{i.e.,}\xspace}

\newcommand{\eg}{\textit{e.g.,}\xspace}

\usepackage[most,skins,theorems]{tcolorbox}
\tcbuselibrary{skins,breakable,fitting,hooks}  
\tcbset{
  aibox/.style={
    width=\linewidth,
    top=7pt,
    bottom=2pt,
    colback=rliablered!18!white,
    colframe=black,
    colbacktitle=black,
    enhanced,
    center,
    attach boxed title to top left={yshift=-0.1in,xshift=0.15in},
    boxed title style={boxrule=0pt,colframe=white,},
  }
}
\tcbset{
  aibox2/.style={
    width=\linewidth,
    top=7pt,
    bottom=2pt,
    colback=green!18!white,
    colframe=black,
    colbacktitle=black,
    enhanced,
    center,
    attach boxed title to top left={yshift=-0.1in,xshift=0.15in},
    boxed title style={boxrule=0pt,colframe=white,},
  }
}
\newtcolorbox{AIbox}[2][]{aibox,title=#2,#1}
\newtcolorbox{AIbox2}[2][]{aibox2,title=#2,#1}

\hypersetup{
    colorlinks=true,            
    linkcolor=blue,             
    filecolor=magenta,          
    urlcolor=cyan,              
    citecolor=purple,             
    pdftitle={Overleaf Example},
    pdfpagemode=FullScreen,
}

\definecolor{iquestblue}{HTML}{173C7F}
\definecolor{iquestazure}{HTML}{528FCC}

\newcommandx{\info}[2][1=]{\todo[linecolor=red,backgroundcolor=red!25,bordercolor=red,#1]{#2}}

\title{
\vspace{-0.2in}
\centering \fontsize{15pt}{16pt}\selectfont
Towards Effective Experiential Learning: \\
Dual Guidance for Utilization and Internalization
\vspace{-0.2in}
}

\author{
Fei Bai\textsuperscript{1*},
Zhipeng Chen\textsuperscript{1*},
Chuan Hao\textsuperscript{2},
Ming Yang\textsuperscript{2}, \\
\normalfont
Ran Tao\textsuperscript{2},
Bryan Dai\textsuperscript{2},
Wayne Xin Zhao\textsuperscript{1$\dagger$},
Jian Yang\textsuperscript{3},
Hongteng Xu\textsuperscript{1} \\
\normalfont
\textsuperscript{1}Gaoling School of Artificial Intelligence, Renmin University of China, \\
\normalfont
\textsuperscript{2}IQuest Research,
\textsuperscript{3}Beihang University \\
\normalfont
\normalsize{\textsuperscript{*}Equal contributions \quad
\textsuperscript{$\dagger$}Corresponding Author \\
\normalfont
Email: \texttt{feibai@ruc.edu.cn}, \texttt{batmanfly@gmail.com}
}}

\begin{abstract}
Recently, reinforcement learning~(RL) has become an important approach for improving the capabilities of large language models~(LLMs). In particular, reinforcement learning from verifiable rewards~(RLVR) has emerged as a promising paradigm for reasoning tasks. However, existing RL-based training still remains only a rough approximation to human learning. Human learners leverage both external and internal experience to guide exploration and gradually internalize useful trajectories into stable knowledge. Motivated by this gap, we ask: how can LLMs better utilize and internalize experience during RLVR training?
To answer this question, we propose \textbf{D}ual \textbf{G}uidance \textbf{O}ptimization~(\textbf{DGO}), a unified framework that leverages \emph{external} and \emph{internal experience} to improve training effectiveness. Specifically, DGO first constructs an experience bank from previously explored trajectories. The policy then performs exploration under the joint guidance of the experience bank and the model's internal knowledge. The resulting trajectories are further used to refine the experience bank and optimize model parameters, forming a closed loop of experience utilization and internalization. Experiments show that DGO consistently outperforms baseline methods, suggesting that better utilization and internalization of experience lead to more effective reasoning.
\end{abstract}

\begin{document}

\maketitle

\let\oldthefootnote\thefootnote




\section{Introduction}

Recently, reinforcement learning~(RL) has shown great promise in enhancing the performance of large language models (LLMs)~\cite{zhao2023survey,guo2025deepseek}. By treating LLMs as agents, RL encourages them to adjust their behavior towards higher rewards, as in the increasingly popular paradigm of reinforcement learning from verifiable rewards~(RLVR). Through this process, the model learns from correct trajectories while steering away from incorrect ones, thereby progressively improving its reasoning capabilities~\cite{shao2024deepseekmath,deng2025trial}. 


In the literature, various approaches have been proposed to improve RL training performance, among which experiential learning has emerged as an important technique in this line of research~\cite{zhan2025exgrpo,tang2025agent,wang2025cogito}. The concept of experiential learning is actually borrowed from human learners. In real-world scenarios, when human learners exhibit incorrect behaviors, they tend to avoid repeating such actions and gradually accumulate effective policies to achieve better performance, which essentially reflects a form of experience internalization. On the other hand, such experiences can also be acquired from external sources, such as books or even another expert in the task domain~\cite{kolb2014experiential}. Humans naturally develop a dual mechanism for utilizing and internalizing experiences in this manner. Grounded in this learning paradigm, they continuously improve, progressively expanding both their knowledge and problem-solving capabilities~\cite{shirouzu2002cognitively}. 

\begin{wrapfigure}{r}{0.5\textwidth}
\captionsetup{justification=justified,singlelinecheck=false}
  \centering
  \vspace{-10pt}
  \includegraphics[width=0.48\textwidth]{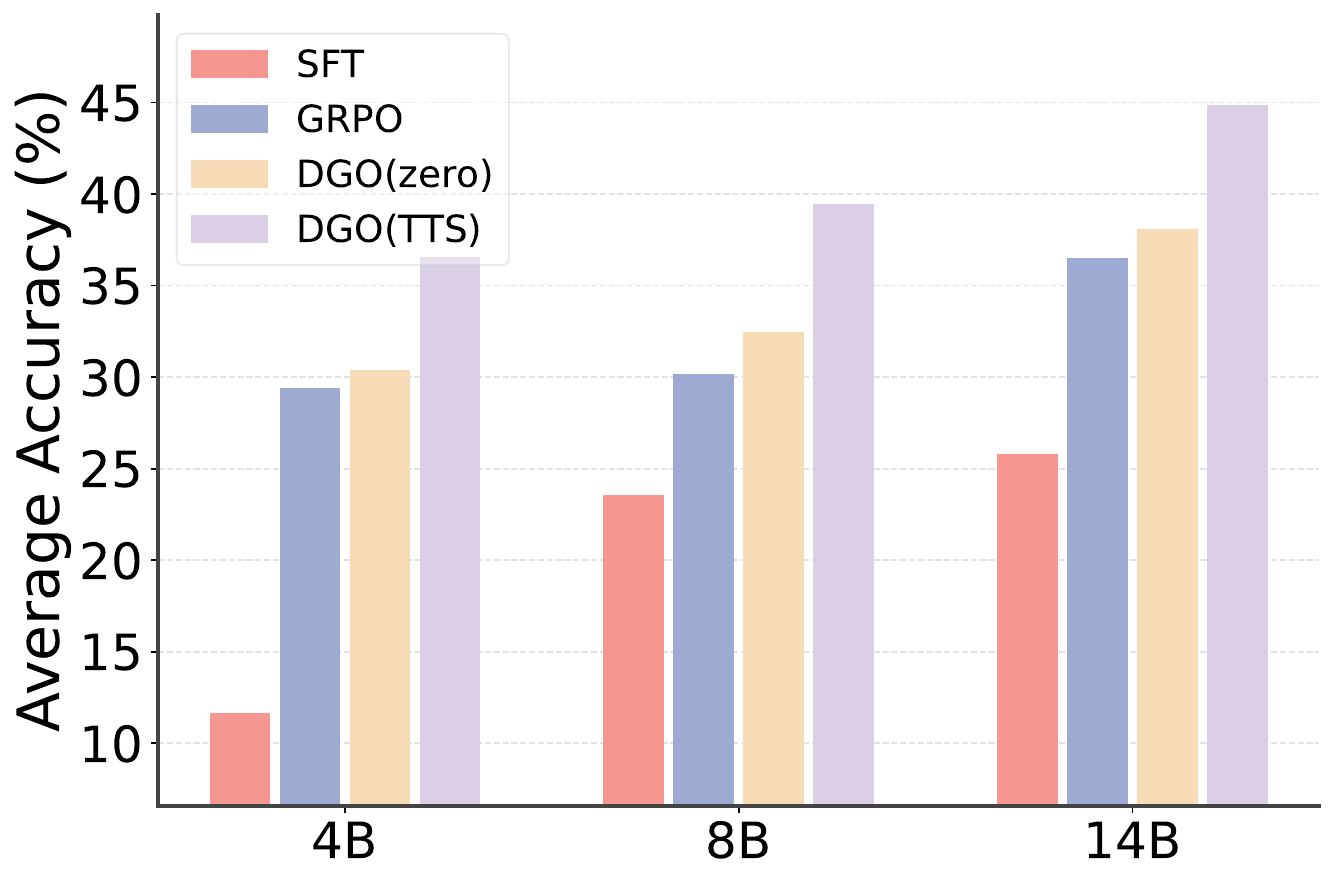}
  \caption{Comparison of DGO and baseline methods on six in-domain and out-of-domain benchmarks using Qwen3-4B/8B/14B-Base.}
  \label{fig:intro-bar}
  \vspace{-10pt}
\end{wrapfigure}


By analogy to human learning, an ideal RL training algorithm should excel at both utilizing and internalizing experience. However, existing efforts tend to focus mainly on one of these two mechanisms. Some approaches emphasize learning from internal knowledge~\cite{ouyang2022training,yu2025dapo,xi2025bapo}, while others focus on leveraging external experience for reasoning~\cite{zhao2024expel,cai2025flex,xu2025mem}. The former places less emphasis on explicit experience utilization and may therefore risk forgetting, whereas the latter relies more on external knowledge and may limit the development of intrinsic capabilities. Therefore, we argue that unifying these two aspects offers a more principled direction for sustained capability improvement over time~\cite{gao2025survey}.



To address this, in this paper, we propose \textbf{D}ual \textbf{G}uidance \textbf{O}ptimization~(\textbf{DGO}), a unified framework that leverages \emph{external} and \emph{internal experience} as guidance to improve training effectiveness. 
DGO formulates model evolution as a coupled process driven by \emph{utilization} and \emph{internalization}: the former guides exploration toward better trajectories based on dual guidance, while the latter leverages these explored trajectories to optimize model parameters. 
Together, they define a closed-loop dynamic between experience-guided exploration and parametric learning.
Concretely, this dynamic unfolds through three iterative stages, \ie experience construction, joint trajectory–policy refinement, and experience internalization. 
First, we utilize an experience generator to construct the experience bank from previously collected trajectories.
Next, the policy explores possible solutions to the given task under the external and internal experience guidance.
Finally, the explored trajectories are rewritten and distilled into the policy parameters to consolidate stable capabilities. 
Repeating this loop progressively improves both experience utilization and internalization across model scales. 
Our contributions can be summarized as follows,

\begin{itemize}


    \item We show that effective experience utilization and internalization improve LLM reasoning and broaden the range of reasoning behaviors the model can reliably reach. In Section~\ref{sec:main_results}, we observe that reasoning performance improves as training iterations and reasoning rounds increase.
    
    \item We introduce \textbf{Dual Guidance Optimization (DGO)}, a unified framework that formalizes the interaction between parametric training and non-parametric experience, offering a new perspective on the model learning paradigm.

    \item DGO achieves the best average score of 32.41\% across six challenging benchmarks on Qwen3-8B-Base under intrinsic inference, and test-time scaling further improves it to 39.38\%, demonstrating the effectiveness of our method.
    
    
\end{itemize}

\section{Related Work}
Experiential learning can be categorized into non-parametric and parametric paradigms. We discuss related work from these two perspectives below.

\paragraph{Non-Parametric Learning.}
Non-parametric learning aims to improve reasoning by enabling models to leverage external guidance, referred to here as experience, such as memory from historical sessions~\cite{chhikara2025mem0,salama2025meminsight}, reusable strategies distilled from generated solutions~\cite{cai2025flex}, and expert demonstrations~\cite{brown2020language}. Existing work mainly falls into two categories: 1) experience construction and retrieval, and 2) experience utilization and internalization. Prior studies have primarily advanced the former through hierarchical experience repositories~\cite{tang2025agent,yang2025reasonflux}, updating mechanisms, and verification strategies~\cite{cao2025remember,kirtania2025improving}. In contrast, the latter remains less explored. While some recent studies have examined either how models can better utilize external experience~\cite{xia2026skillrl} during reasoning or how such experience can be internalized into model parameters~\cite{zhao2026self}, they tend to emphasize only one aspect. Coupling utilization and internalization in a unified framework remains important yet challenging.


\paragraph{Parametric Learning.}
Parametric learning improves LLM reasoning by directly updating model parameters to reinforce effective reasoning behaviors and promote capability internalization. Representative methods include supervised fine-tuning, reinforcement learning~\cite{ouyang2022training,shao2024deepseekmath,yu2025dapo,zheng2025group}, preference optimization~\cite{rafailov2023direct}, and verifier-based training~\cite{lightman2023let}. These approaches have shown strong effectiveness in improving reasoning performance through explicit parameter optimization. However, they typically treat high-quality trajectories primarily as training signals for parameter updates, rather than reusable experience that can be retained and leveraged beyond the current optimization step~\cite{didolkar2025metacognitive}. As a result, valuable reasoning trajectories are often not explicitly reused, limiting their potential to provide sustained guidance for future problem solving~\cite{luo2025empirical}.


\begin{table}[t]
\centering
\small
\setlength{\tabcolsep}{4pt}
\begin{tabular}{lcccc}
\toprule
\multirow{2}{*}{\textbf{Method}} & \multicolumn{2}{c}{\textbf{Optimization Paradigm}} & \multicolumn{2}{c}{\textbf{Enhanced Capability}} \\
\cmidrule(lr){2-3} \cmidrule(lr){4-5}
 & Non-parameter & Parameter & Utilization & Internalization \\
\midrule
SFT                 & \ding{55} & \ding{51} & \ding{55} & \ding{51} \\
GRPO                & \ding{55} & \ding{51} & \ding{55} & \ding{51} \\
On-Policy Self-Distillation                & \ding{51} & \ding{51} & \ding{55} & \ding{51} \\
Skill-Augmented RL             & \ding{51} & \ding{51} & \ding{51} & \ding{55} \\
\textbf{DGO (Ours)} & \ding{51} & \ding{51} & \ding{51} & \ding{51} \\
\bottomrule
\end{tabular}
\caption{Comparison of representative methods in terms of optimization form and their primary emphasis on experience utilization and internalization. DGO integrates parametric and non-parametric optimization within a unified framework that considers both aspects.}
\label{tab:intro_comparison_final}
\end{table}
\section{Method}
In this section, we present the details of DGO. For clarity, Table~\ref{tab:intro_comparison_final} provides a conceptual positioning of DGO relative to representative methods.

\begin{figure*}[h]
\captionsetup{justification=justified,singlelinecheck=false}
  \centering
  \includegraphics[width=1.0\linewidth]{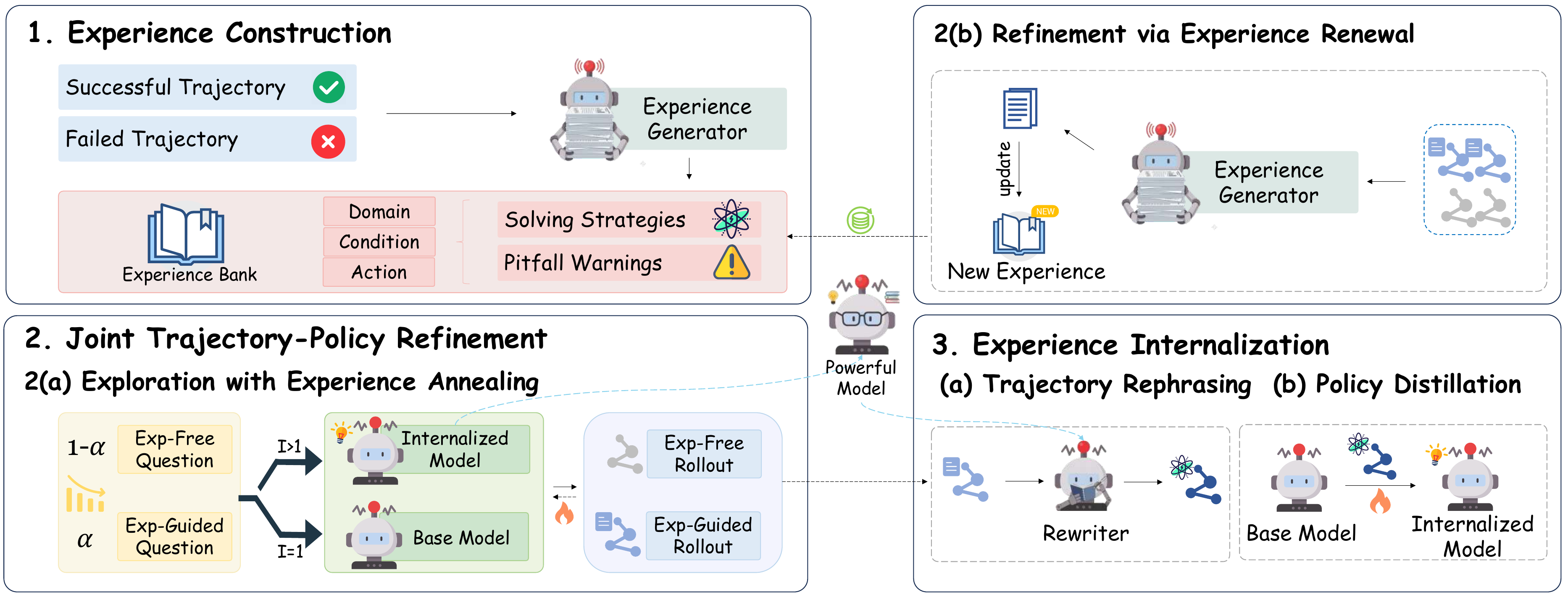}
  \caption{Our framework consists of three components: 1) Experience Construction, 2) Joint Trajectory–Policy Refinement, and 3) Experience Internalization, forming a closed-loop process that enhances experience utilization and internalization. $I$ denotes the training iteration, and the final model is the RL policy from the last iteration.}
  \label{fig:main}
\end{figure*}

\subsection{Overview of Dual Guidance Optimization}
\label{sec:4.1}

Dual Guidance Optimization (DGO) is based on a simple observation: reasoning can be improved by two complementary sources of guidance. External guidance helps the model explore solution trajectories that are difficult to discover under the current policy alone, while internal guidance, encoded in the model parameters, supports the model's own reasoning capability. DGO links the two by transforming trajectories discovered under external guidance into parameter updates, allowing useful reasoning patterns to be progressively internalized.

Formally, given an input problem $x$, an experience bank $\mathcal{E}$, and a policy $\pi_\theta$, the model generates trajectories under the joint effect of external and internal guidance as
\begin{equation}
s \sim \pi_\theta(\cdot \mid x, \mathcal{E}).
\end{equation}
The resulting trajectories are then transformed into supervision for parameter update:
\begin{equation}
\theta' \leftarrow \mathrm{Update}(\theta; \mathcal{T}(s)),
\end{equation}
where $\mathcal{T}(\cdot)$ denotes the transformation from generated trajectories to training signals, such as filtering and rewriting. Through this process, external and internal guidance form a closed feedback loop in which external experience improves trajectory discovery, and the resulting trajectories in turn refine experience, strengthen the model's internal guidance, and support more effective use of external experience.



\subsection{Iterative Dual-Guidance Training}

\begin{algorithm}[t]
\small
\caption{Dual Guidance Optimization}
\label{alg:dgs}
\begin{algorithmic}[1]
    \State \textbf{Input:} Training set $\mathcal{D}$, initial model $\pi_{\theta^{(0)}}$, pre-built experience bank $\mathcal{E}$, experience generator $G$, number of stages $K$, annealing schedule $\{\alpha_k\}_{k=1}^K$, internalization ratios $\{\beta_k\}_{k=1}^K$
    \State \textbf{Output:} Final policy $\pi_{\theta_{\mathrm{RL}}^{(K)}}$
    
    \State $\theta_{\mathrm{start}}^{(1)} \gets \theta^{(0)}$
    \State $\mathcal{S}_{\mathrm{imp}}^{(\le 0)} \gets \emptyset$, $\mathcal{S}_{e}^{(\le 0)} \gets \emptyset$
    
    \For{$k = 1$ to $K$}
        \State \textcolor{gray}{\textit{// Joint Trajectory-Policy Refinement}}
        \State Retrieve relevant experience from $\mathcal{E}$ and construct $\mathcal{D}_{e}^{(k)}$
        \State Construct mixed data with annealing coefficient $\alpha_k$
        \State $\mathcal{D}_{\mathrm{RL}}^{(k)} \gets
        \alpha_k \mathcal{D}_{e}^{(k)} + (1-\alpha_k)\mathcal{D}_{\mathrm{free}}^{(k)}$
        \State $\theta_{\mathrm{RL}}^{(k)} \gets
        \mathrm{Update}(\theta_{\mathrm{start}}^{(k)}, \mathcal{D}_{\mathrm{RL}}^{(k)})$
        \State Collect rollout trajectories $\mathcal{S}^{(k)}$
        \State Update the experience bank $\mathcal{E}$ using $\mathcal{S}^{(k)}$ and $G$
        
        \State \textcolor{gray}{\textit{// Experience Internalization}}
        \State $\widetilde{\mathcal{S}}_{e}^{(k)} \gets
        \mathrm{Rewrite}(\mathcal{S}_{e}^{(k)})$
        \State $\mathcal{S}_{\mathrm{imp}}^{(k)} \gets
        \mathrm{Filter}(\widetilde{\mathcal{S}}_{e}^{(k)})$
        \State $\mathcal{S}_{\mathrm{imp}}^{(\le k)} \gets
        \mathcal{S}_{\mathrm{imp}}^{(\le k-1)} \cup \mathcal{S}_{\mathrm{imp}}^{(k)}$
        \State $\mathcal{S}_{e}^{(\le k)} \gets
        \mathcal{S}_{e}^{(\le k-1)} \cup \mathcal{S}_{e}^{(k)}$
        \State $\mathcal{D}_{\mathrm{dist}}^{(k)} \gets
        \beta_k \mathcal{S}_{\mathrm{imp}}^{(\le k)} +
        (1-\beta_k)\mathcal{S}_{e}^{(\le k)}$
        \State $\theta_{\mathrm{start}}^{(k+1)} \gets
        \mathrm{Distill}(\theta^{(0)}, \mathcal{D}_{\mathrm{dist}}^{(k)})$
        \State \textcolor{gray}{\textit{// used to warm-start the next RL iteration}}
    \EndFor
    
    \State \textbf{return} $\pi_{\theta_{\mathrm{RL}}^{(K)}}$
\end{algorithmic}
\end{algorithm}

Building on Section~\ref{sec:4.1}, we implement an iterative training cycle, illustrated in Figure~\ref{fig:main}, consisting of experience construction, joint trajectory-policy refinement, and experience internalization. The pseudocode is shown in Algorithm~\ref{alg:dgs}.

\subsubsection{Experience Construction}

Experience is distilled from historical reasoning trajectories to provide external guidance for future problem solving. Rather than directly reusing full reasoning trajectories, we distill transferable strategies and diagnostic cues that can generalize across problems. Each experience is represented as a triplet $e = (c, \kappa, \tau)$,
where $c$ specifies the broader domain or problem phase, $\kappa$ identifies the key condition, scenario, or potential error, and $\tau$ prescribes the corresponding action to take or avoid. 

Given a trajectory $s$, a unified experience generator $G$ extracts an experience as $e = G(s)$. Experiences derived from correct trajectories capture effective solving strategies, while those derived from incorrect trajectories capture pitfall warnings. Collectively, these experiences form the bank $\mathcal{E} = \{ e_i \}_{i=1}^{N}$. To ensure that the bank provides transferable reasoning guidance rather than answer-specific supervision, any experience that contains or strongly suggests the final answer is strictly filtered out. The extraction prompt is shown in Table~\ref{tab:experience_prompt}, and examples are provided in Table~\ref{tab:experience_example}.

\subsubsection{Joint Trajectory-Policy Refinement}
In this stage, we employ RL (\eg GRPO) to strengthen experience-guided reasoning while continuously producing higher-quality trajectories for subsequent experience internalization.


\paragraph{Exploration with Experience Annealing.}
We use RL to strengthen the model’s ability to utilize external experience while gradually reducing its dependence on such guidance. To this end, at iteration $k$, training draws instances from a mixture of the experience-guided distribution $D_e^{(k)}$ and the experience-free distribution $D_i^{(k)}$, where the annealing coefficient $\alpha_k$ controls the strength of experience guidance\footnote{$\alpha_k \mathcal{D}_e^{(k)} + (1-\alpha_k)\mathcal{D}_i^{(k)}$ denotes a mixed dataset constructed by sampling examples from $\mathcal{D}_e^{(k)}$ and $\mathcal{D}_i^{(k)}$ with probabilities $\alpha_k$ and $1-\alpha_k$, respectively, rather than a literal weighted sum of sets. We use this notation throughout the paper.}:
\begin{equation}
D^{(k)} = \alpha_k D_e^{(k)} + (1 - \alpha_k) D_i^{(k)}.
\end{equation}
As training progresses, $\alpha_k$ gradually decreases, so that the policy relies more heavily on external experience in earlier stages for guided exploration, and progressively shifts toward weaker guidance in later stages. Let $z=(x,\tilde{e})$ denote a training instance, where $\tilde{e}$ is the retrieved experience for experience-guided instances and is empty otherwise. The policy is optimized over the mixed training distribution using
\begin{equation}
\begin{aligned}
\mathcal{L}(\theta) =\ 
& \mathbb{E}_{z \sim D^{(k)},\, s \sim \pi_{\theta_{\text{old}}}(\cdot \mid z)} \Bigg[
\frac{1}{|s|} \sum_{t=1}^{|s|} \min \Big( \rho_t(\theta) A_t, \\
&\quad \operatorname{clip}(\rho_t(\theta), 1 - \epsilon, 1 + \epsilon) A_t \Big)
\Bigg],
\label{equation:obj}
\end{aligned}
\end{equation}
where
$\rho_t(\theta)=
\frac{\pi_{\theta}(s_t \mid z, s_{<t})}
{\pi_{\theta_{\text{old}}}(s_t \mid z, s_{<t})}$.
This annealing process encourages the model to benefit from strong experience guidance early on, while gradually reducing reliance on external experience without losing the ability to use it effectively.

\paragraph{Refinement via Experience Renewal.}
After each RL iteration $k$, a set of rollout trajectories
\(
\mathcal{S}^{(k)} = \{ s_j^{(k)} \}
\)
is collected. The experience generator $G$ summarizes these trajectories into a new batch of experience:
$\mathcal{E}_{\text{new}}^{(k)} = G\!\left(\mathcal{S}^{(k)}\right)$,
which is merged with the current bank and filtered through quality-based selection as shown in Appendix~\ref{Exp Update}:
\begin{equation}
\mathcal{E}^{(k+1)} 
= \operatorname{Select}\!\left(
\mathcal{E}^{(k)} \cup \mathcal{E}_{\text{new}}^{(k)}
\right),
\end{equation}
where $\operatorname{Select}(\cdot)$ retains the most informative experiences for the next iteration. The updated bank $\mathcal{E}^{(k+1)}$ is then used to construct the mixed training distribution $D^{(k+1)}$ for later policy refinement.

As trajectory quality improves across iterations, the extracted experience becomes increasingly useful for guiding later exploration. This creates a positive feedback loop: improved trajectories yield higher-quality experience, which in turn supports further trajectory refinement and provides better data for subsequent experience internalization.

\subsubsection{Experience Internalization}
\label{sec:4.2.3}

Experience internalization transfers the reasoning gains induced by external experience into the policy parameters through \emph{trajectory rephrasing} and \emph{policy distillation}. In this way, the model gradually absorbs the useful reasoning patterns encouraged by external experience, improving its standalone reasoning ability even without explicit experience.

\paragraph{Trajectory Rephrasing.}
External experience can guide the model to produce trajectories that would be difficult to obtain through unguided reasoning alone. However, such trajectories often contain explicit references to the provided experience, which can introduce undesirable dependencies if used directly for training. To address this, we rewrite experience-guided trajectories into self-contained reasoning traces that remove explicit references while preserving their underlying rationale.

Given a problem $x$ with retrieved experience $\mathcal{E}_x$ and an experience-guided trajectory $s_e^{(k)} \in \mathcal{S}_e^{(k)}$, we use the RL checkpoint $\theta_{\mathrm{RL}}^{(k)}$ with a rewriting prompt template $p$~(see in Table~\ref{tab:rewrite_prompt}) to generate a rewritten trajectory:
\begin{equation}
\tilde{s}_e^{(k)} 
\sim 
\pi_{\theta_{\mathrm{RL}}^{(k)}}
\!\left(
\cdot \mid x, s_e^{(k)}, \mathcal{E}_x, p
\right).
\label{eq:rewrite_gen}
\end{equation}

Applying this procedure to all trajectories in $\mathcal{S}_e^{(k)}$ yields the rewritten trajectory set $\tilde{\mathcal{S}}_e^{(k)}$. A rewritten trajectory is retained only if it satisfies strict filtering constraints, including correctness validation, keyword exclusion, and length control, with detailed settings provided in Appendix~\ref{app:filter}. Denoting the filtering operator by $\mathcal{F}(\cdot)$, the implicit trajectory set at iteration $k$ is defined as
\begin{equation}
\mathcal{S}_{\mathrm{imp}}^{(k)}
=
\operatorname{Top\text{-}K}
\Big(
\mathcal{F}\big(
\tilde{\mathcal{S}}_e^{(k)}
\big)
\Big),
\label{eq:rewrite_filter}
\end{equation}
where up to $K$ solutions are retained per problem.

\paragraph{Policy Distillation.}

We further internalize external experience into the model parameters through supervised distillation over selected trajectories. At iteration $k$, we accumulate both implicit and experience-guided trajectories as
$\mathcal{S}_{\mathrm{imp}}^{(\le k)}=\bigcup_{i=1}^{k}\mathcal{S}_{\mathrm{imp}}^{(i)}$
and
$\mathcal{S}_{e}^{(\le k)}=\bigcup_{i=1}^{k}\mathcal{S}_{e}^{(i)}$,
respectively.
To preserve the model’s ability to utilize external experience, we construct the distillation dataset by mixing these two trajectory sets:
\begin{equation}
\mathcal{D}_{\mathrm{dist}}^{(k)}
=
\beta_k \,
\mathcal{S}_{\mathrm{imp}}^{(\le k)}
+
(1-\beta_k)\,
\mathcal{S}_{e}^{(\le k)},
\label{eq:dist_data_k}
\end{equation}
where $\beta_k \in [0,1]$ controls the mixing ratio. The policy is then optimized by maximum likelihood, with parameters initialized from the initial checkpoint $\theta^{(0)}$ to mitigate overfitting:
\begin{equation}
\theta^{(k+1)} 
=
\arg\max_{\theta}
\;
\mathbb{E}_{(x,s)\sim\mathcal{D}_{\mathrm{dist}}^{(k)}}
\log \pi_{\theta}(s \mid x).
\label{eq:rft_obj}
\end{equation}
The resulting model, denoted as the \emph{Internalized Model}, is then used to initialize the next iteration.

\subsection{Inference Paradigm}

\paragraph{Intrinsic Inference.}
This mode evaluates the model’s reasoning ability without external guidance. Given a problem $x$, the model generates a trajectory solely from its parametric knowledge:
\begin{equation}
s \sim \pi_\theta(\cdot \mid x).
\end{equation}

\paragraph{Iterative Experience-Guided Test-Time Scaling.}
Given a problem $x$, the model performs $K$ rounds of trajectory generation. In each round $k$, it produces $N$ trajectories, from which the experience generator $G$ extracts online experience $\mathcal{E}_{\text{online}}^{(k)}$ for the next round. The first round uses retrieved offline experience $\mathcal{E}_{\text{offline}}$ to provide a cold start, while later rounds use the online experience extracted from the previous round\footnote{The experience generator does not access correctness labels and must infer correctness before summarizing experience; see Table~\ref{tab:experience_prompt}.}:
\begin{equation}
s^{(k)} \sim
\begin{cases}
\pi_\theta(\cdot \mid x, \mathcal{E}_{\text{offline}}), & k = 1,\\[2mm]
\pi_\theta(\cdot \mid x, \mathcal{E}_{\text{online}}^{(k-1)}), & k > 1.
\end{cases}
\label{eq:iegi_generation}
\end{equation}
\section{Experiments}
In this section, we first detail the experimental setup and then
report the results and detailed analysis.

\begin{table*}
\captionsetup{justification=justified,singlelinecheck=false}
\centering
\resizebox{\linewidth}{!}{
\begin{tabular}{llccccccc}
\toprule
\multirow{2}{*}{\textbf{Model}} 
& \multirow{2}{*}{\textbf{Method}} 
& \multicolumn{4}{c}{\textbf{In-domain~(\%)}} 
& \multicolumn{2}{c}{\textbf{Out-of-domain~(\%)}} 
& \multirow{2}{*}{\textbf{Avg}} \\

\cmidrule(lr){3-6} \cmidrule(lr){7-8}
& 
& AIME24 
& AIME25 
& HMMT25 
& MATH500 
& GPQA-D 
& HLE 
&  \\

\midrule

\multirow{6}{*}{\textbf{Qwen3-4B-Base}}

& EGI
& 1.46 & 1.46 & 0.00 & 23.52 & 12.12 & 1.31 & 6.64 \\

& SFT
& 5.83 & 5.62 & 1.04 & 31.52 & 22.92 & 2.56 & 11.58 \\

& GRPO 
& \underline{26.88} & \underline{22.50} & 8.75 & \textbf{72.75} & \underline{42.55} & \underline{2.73} & \underline{29.36} \\

& DAPO 
& 26.04 & 22.29 & \underline{9.17} & 71.35 
& 41.79 & 2.34 & 28.83 \\

& DGO~(zero) 
& \textbf{27.29} & \textbf{23.54} & \textbf{11.67} & \underline{72.60} 
& \textbf{43.56} & \textbf{3.37} & \textbf{30.34} \\

& \cellcolor{lightblue}DGO~(TTS)
& \cellcolor{lightblue}40.21 & \cellcolor{lightblue}27.71 & \cellcolor{lightblue}15.83 & \cellcolor{lightblue}78.35 
& \cellcolor{lightblue}52.53 & \cellcolor{lightblue}4.45 & \cellcolor{lightblue}36.51 \\

\midrule

\multirow{6}{*}{\textbf{Qwen3-8B-Base}}
& EGI
& 6.04 & 4.79 & 2.08 & 41.50 & 22.22 & 3.12 & 13.29 \\

& SFT 
& 15.00 & 11.46 & 3.54 & 70.38 & 36.30 & \underline{4.34} & 23.50 \\

& GRPO 
& \underline{29.79} & \underline{20.42} & \underline{9.17} & 73.20 & 44.32 & 3.95 & 30.14 \\

& DAPO
& 25.00 & 19.17 & \underline{9.17} & \textbf{80.77} & \underline{45.83} & \textbf{4.66} & \underline{30.77} \\

& DGO~(zero) 
& \textbf{30.21} & \textbf{23.33} & \textbf{11.88} & \underline{75.25} 
& \textbf{49.75} & 4.02 & \textbf{32.41} \\

& \cellcolor{lightblue}DGO~(TTS)
& \cellcolor{lightblue}49.38 & \cellcolor{lightblue}30.63 & \cellcolor{lightblue}14.37 & \cellcolor{lightblue}80.42 
& \cellcolor{lightblue}54.99 & \cellcolor{lightblue}6.51 & \cellcolor{lightblue}39.38 \\

\midrule

\multirow{6}{*}{\textbf{Qwen3-14B-Base}}
& EGI
& 8.33 & 5.42 & 1.67 & 47.02 & 28.91 & 3.29 & 15.77 \\

& SFT 
& 18.12 & 13.33 & 6.46 & 74.58 & 37.12 & 4.96 & 25.76 \\

& GRPO 
& \underline{38.12} & \underline{27.08} & \underline{14.37} & 89.00 & 45.01 & 5.15 & 36.45 \\

& DAPO
& 35.83 & 26.25 & 14.17 & \underline{89.50} & \underline{49.37} & \underline{5.45} & \underline{36.76} \\

& DGO~(zero)
& \textbf{38.54} & \textbf{29.17} & \textbf{15.21} & \textbf{90.60} 
& \textbf{50.13} & \textbf{5.63} & \textbf{38.04} \\

& \cellcolor{lightblue}DGO~(TTS)
& \cellcolor{lightblue}54.48 & \cellcolor{lightblue}39.38 & \cellcolor{lightblue}22.29 & \cellcolor{lightblue}93.17 
& \cellcolor{lightblue}51.64 & \cellcolor{lightblue}7.71 & \cellcolor{lightblue}44.78 \\

\bottomrule
\end{tabular}
}
\caption{Performance comparison of our method with other approaches across three model scales on both in-domain and out-of-domain benchmarks. \textbf{Bold} indicates the best result within each model, and \underline{underlined} indicates the second-best. DGO~(zero) and DGO~(TTS) denote intrinsic inference and test-time scaling, respectively.} 
\label{tab:main_1}
\end{table*}

\subsection{Experiment Settings}

\paragraph{Training.}
We use Qwen3 models~\cite{yang2025qwen3} at three scales~(4B, 8B, and 14B) as policy backbones, and sample 9,600 questions from SkyWork~\cite{skywork-or1-2025} for RL training. Training consists of three GRPO-based iterations with iterative experience update and ratio annealing: the experience ratio $\alpha$ decreases from 1.0 to 0.5 and 0.25, while the internalization ratio $\beta_k$ is fixed at 0.5. After each iteration, selected trajectories are rewritten into structured experience and distilled into the policy. For the experience generator, we distill 40,000 samples from DeepSeek v3.2~\cite{liu2025deepseek} and finetune Qwen3-8B-Base, which is used throughout training. Details are in Appendix~\ref{training_details}.

\paragraph{Evaluation.}
We evaluate our method on in-domain benchmarks, including AIME24, AIME25, HMMT25-Feb, and MATH500~\cite{hendrycks2021measuringmathematicalproblemsolving}, and out-of-domain benchmarks, including GPQA-Diamond~\cite{rein2024gpqa} and HLE-Verified~\cite{zhai2026hle}. Unless otherwise specified, we use temperature 1.0, top-$p$ 1.0, a maximum sequence length of 16K, and report avg@16. We consider two settings: intrinsic inference, which measures standalone reasoning, and Test-Time Scaling (TTS)~\cite{hu2026pacore}, which evaluates experience-guided inference. For TTS, we use at most $K=5$ iterations, summarize up to $N=4$ experiences per iteration, and retrieve experience by dense vector similarity with all-MiniLM-L6-v2.


\paragraph{Baselines.}
We compare our method with both training-free and training-based baselines. As a representative training-free baseline, we consider simple Experience-Guided Inference (EGI), which conditions the model on external experience at inference time without updating its parameters. For training-based baselines, we include SFT, GRPO~\cite{shao2024deepseekmath}, and DAPO~\cite{yu2025dapo}, which improve reasoning performance through parameter optimization alone, without incorporating external experience during training.

\subsection{Main Results}
\label{sec:main_results}


\paragraph{DGO consistently enhances intrinsic reasoning across model scales.}
As shown in Table~\ref{tab:main_1}, DGO consistently outperforms strong baselines, including SFT and RL-based methods, achieving the best average scores of 30.34\% for 4B, 32.41\% for 8B, and 38.04\% for 14B, with improvements on both challenging mathematical benchmarks, including AIME25 and HMMT25, and broader-coverage benchmarks like MATH500. These results suggest that more effective internalization contributes to stronger intrinsic reasoning in DGO.

\begin{figure}[t]
  \centering
  \begin{subfigure}[b]{0.3\linewidth}
    \centering
    \includegraphics[width=\linewidth]{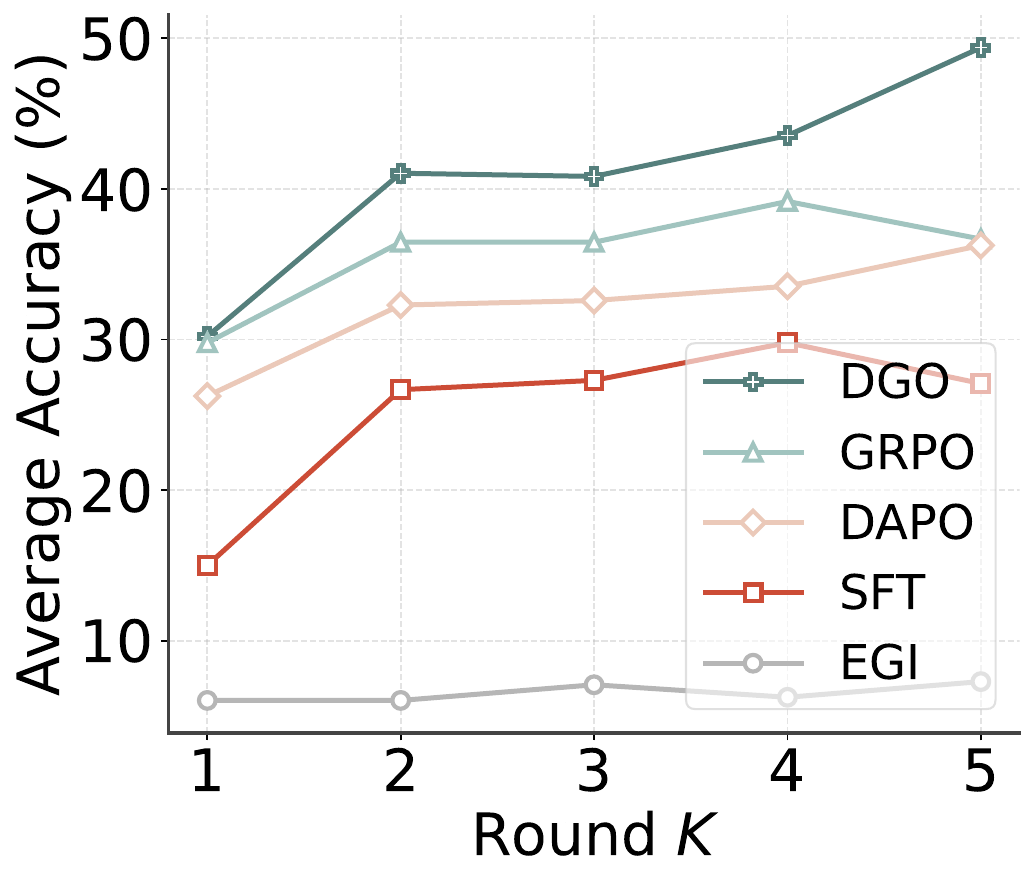}
    \caption{AIME24}
    \label{fig:TTS_k_24}
  \end{subfigure}
  \hspace{0.03\linewidth}
  \begin{subfigure}[b]{0.3\linewidth}
    \centering
    \includegraphics[width=\linewidth]{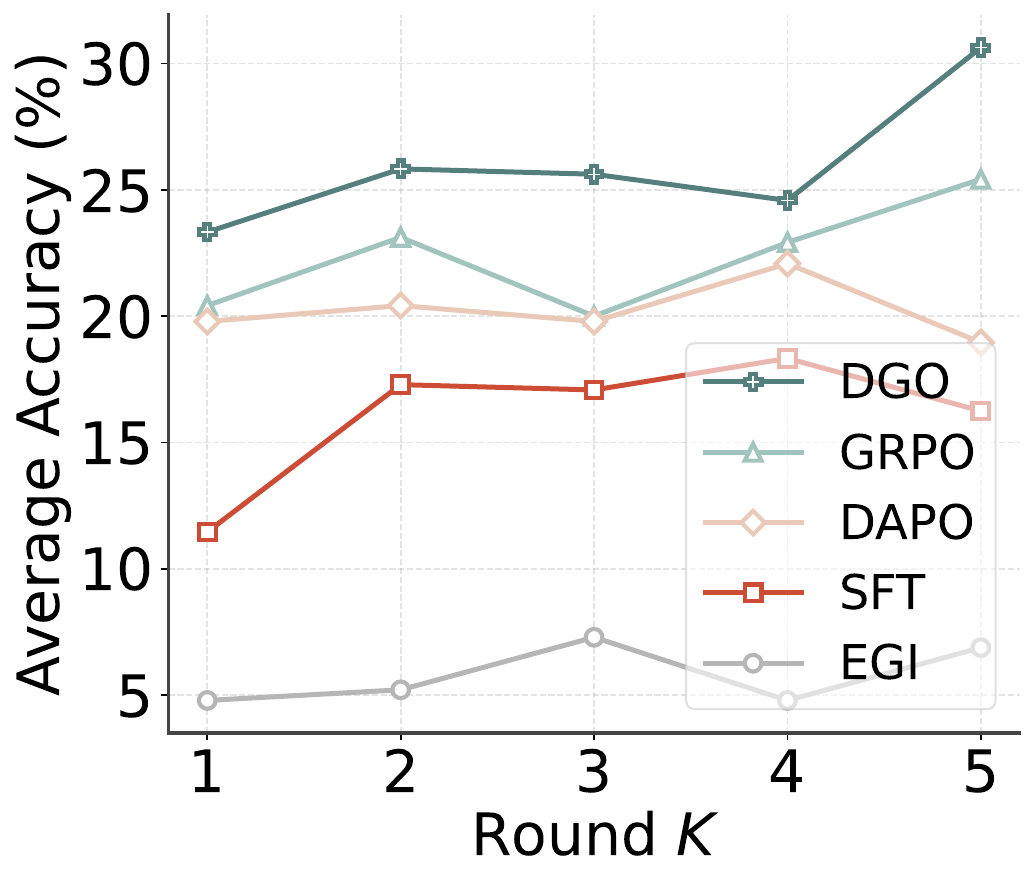}
    \caption{AIME25}
    \label{fig:TTS_k_25}
  \end{subfigure}
  \caption{Accuracy of Qwen3-8B-Base across successive TTS rounds on AIME24 and AIME25, compared with DGO and other baseline methods.}
  \label{fig:TTS_k}
\end{figure}

\paragraph{DGO enables more effective use of experience.}
Incorporating TTS leads to substantial gains over zero-shot inference. For example, on the 4B model the average score rises from 30.34\% to 36.51\%, and on the 14B model from 38.04\% to 44.78\%, indicating that DGO-trained models can effectively benefit from test-time experience. Moreover, as shown in Figure~\ref{fig:TTS_k}, performance improves steadily with successive TTS rounds and consistently surpasses other methods. A more detailed analysis of experience utilization is provided in Section~\ref{utilization}.

\paragraph{DGO shows cross-domain generalization.}
Although trained exclusively on mathematical data, DGO achieves competitive performance on knowledge-intensive benchmarks, including GPQA-Diamond and HLE-Verified. Both DGO~(zero) and DGO~(TTS) perform well on these tasks, suggesting two complementary effects. First, reasoning experience internalized in the mathematical domain can support broader problem-solving. Second, the ability to leverage prior reasoning experience might also transfer across domains.

\begin{table*}[h]
\captionsetup{justification=justified,singlelinecheck=false}
\centering
\small
\setlength{\tabcolsep}{5pt}
\resizebox{0.9\linewidth}{!}{%
\begin{tabular}{lccccccc}
\toprule
\multirow{2}{*}{\textbf{Method}} 
& \multicolumn{4}{c}{\textbf{In-domain~(\%)}} 
& \multicolumn{2}{c}{\textbf{Out-of-domain~(\%)}} 
& \multirow{2}{*}{\textbf{Avg}} \\

\cmidrule(lr){2-5} \cmidrule(lr){6-7}
& AIME24 & AIME25 & HMMT25 & MATH500 & GPQA-D & HLE &  \\

\midrule

DGO~(zero) & \textbf{27.08} & \textbf{21.04} & \textbf{9.17} & \textbf{70.08} & \textbf{40.47} & \underline{2.99} & \textbf{28.47} \\
\multicolumn{1}{r}{w/o \emph{EA}} & 19.38 & 17.50 & 7.71 & 68.08 & 38.64 & \textbf{3.24} & 25.76 \\
\multicolumn{1}{r}{w/o \emph{ER}} & 19.58 & 18.75 & 7.50 & 68.53 & 38.57 & 2.90 & 25.97 \\
\multicolumn{1}{r}{w/o \emph{TR}} & 22.08 & 17.50 & 6.04 & 66.65 & 39.90 & 2.47 & 25.77 \\
\multicolumn{1}{r}{w/o \emph{PD}} & 23.75 & 20.83 & 7.71 & 67.73 & 38.26 & 2.79 & 26.84 \\

\midrule

DGO~(TTS) & \underline{40.83} & \textbf{26.67} & \textbf{15.42} & \textbf{73.90} & \textbf{52.65} & \textbf{4.58} & \textbf{35.68} \\
\multicolumn{1}{r}{w/o \emph{EA}} & 37.71 & 24.58 & 13.12 & 70.43 & 48.30 & 3.37 & 32.92 \\
\multicolumn{1}{r}{w/o \emph{ER}} & \textbf{42.92} & 23.75 & 15.21 & 70.78 & 48.42 & 3.11 & 34.03 \\
\multicolumn{1}{r}{w/o \emph{TR}} & 31.87 & 20.42 & 11.67 & 72.10 & 48.11 & 3.99 & 31.36 \\
\multicolumn{1}{r}{w/o \emph{PD}} & 34.58 & 22.71 & 15.00 & 67.00 & 49.31 & 3.20 & 31.97 \\

\bottomrule
\end{tabular}%
}
\caption{Ablation study on Qwen-4B-Base. \emph{EA}, \emph{ER}, \emph{TR}, and \emph{PD} denote \emph{Experience Annealing}, \emph{Experience Renewal}, \emph{Trajectory Rephrasing}, and \emph{Policy Distillation} respectively. Experiments are conducted between iteration 1 and 2.}
\label{tab:abla}
\end{table*}

\subsection{Ablation Studies}
In this section, we conduct ablation studies to evaluate the contribution of key design choices in DGO.

\subsubsection{Component Ablation}
We conduct component-level ablations to assess the contribution of each component in DGO. As shown in Table~\ref{tab:abla}, removing any of the four components consistently hurts performance, indicating that each is important to the overall effectiveness.

$\bullet$ \emph{Experience Renewal} keeps the experience pool informative and diverse throughout training. Removing it lowers the average score from 28.47\% to 25.97\% for DGO~(zero) and from 35.68\% to 34.03\% for DGO~(TTS), showing that maintaining fresh experience supports stronger subsequent trajectories and more effective internalization.

$\bullet$ \emph{Trajectory Rephrasing} improves training trajectories by removing noisy patterns that explicitly copy experience, making them easier to internalize. Without it, the average score drops from 28.47\% to 25.77\% for DGO~(zero) and from 35.68\% to 31.36\% for DGO~(TTS). This also suggests that effective trajectory cleanup can be achieved by the model itself, without requiring a stronger external rewriter.

$\bullet$ \emph{Experience Annealing} and \emph{Policy Distillation} are crucial for turning external experience into lasting parametric gains. Annealing reduces over-reliance on explicit experience and encourages the model to absorb underlying reasoning patterns, while distillation consolidates these gains into the policy. Removing annealing drops the average score from 28.47\% to 25.76\% for DGO~(zero) and from 35.68\% to 32.92\% for DGO~(TTS), and removing distillation also causes clear degradation.


\subsubsection{Role of Iterative Training}
\begin{figure}[t]
  \centering
  \begin{subfigure}[b]{0.3\linewidth}
    \centering
    \includegraphics[width=\linewidth]{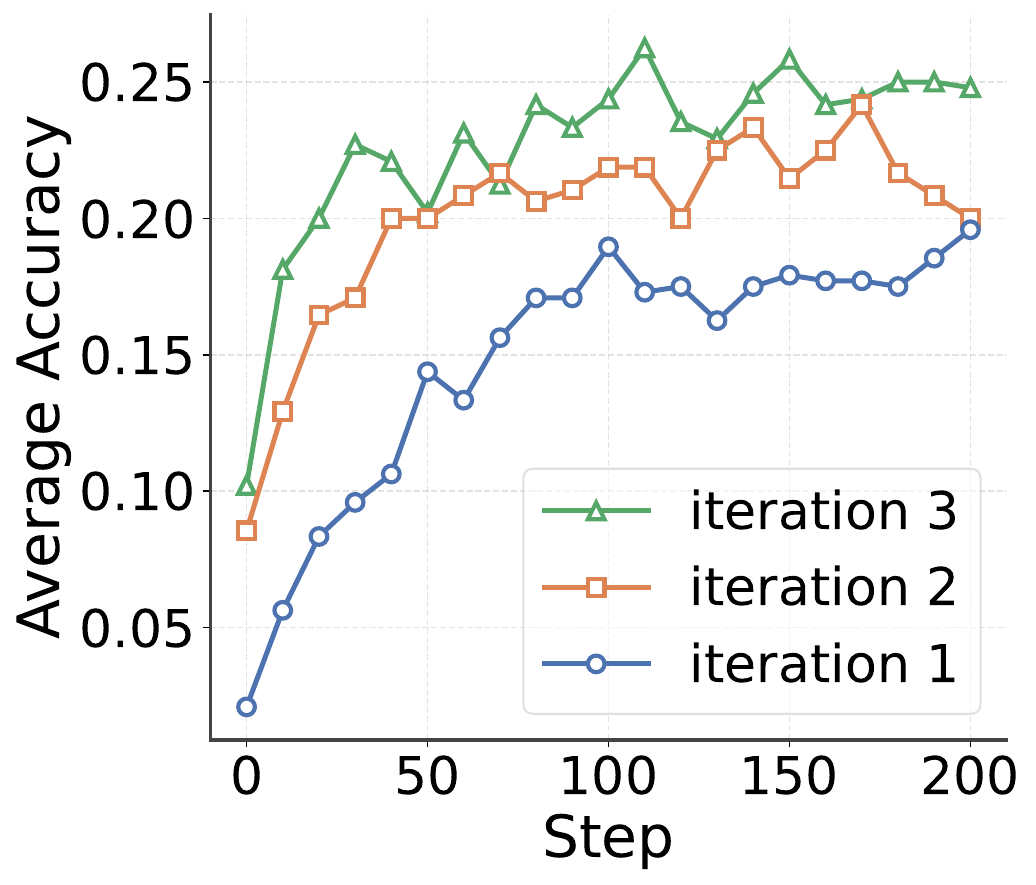}
    \caption{Qwen3-8B-Base}
    \label{fig:iter_8b}
  \end{subfigure}
  \hspace{0.03\linewidth}
  \begin{subfigure}[b]{0.3\linewidth}
    \centering
    \includegraphics[width=\linewidth]{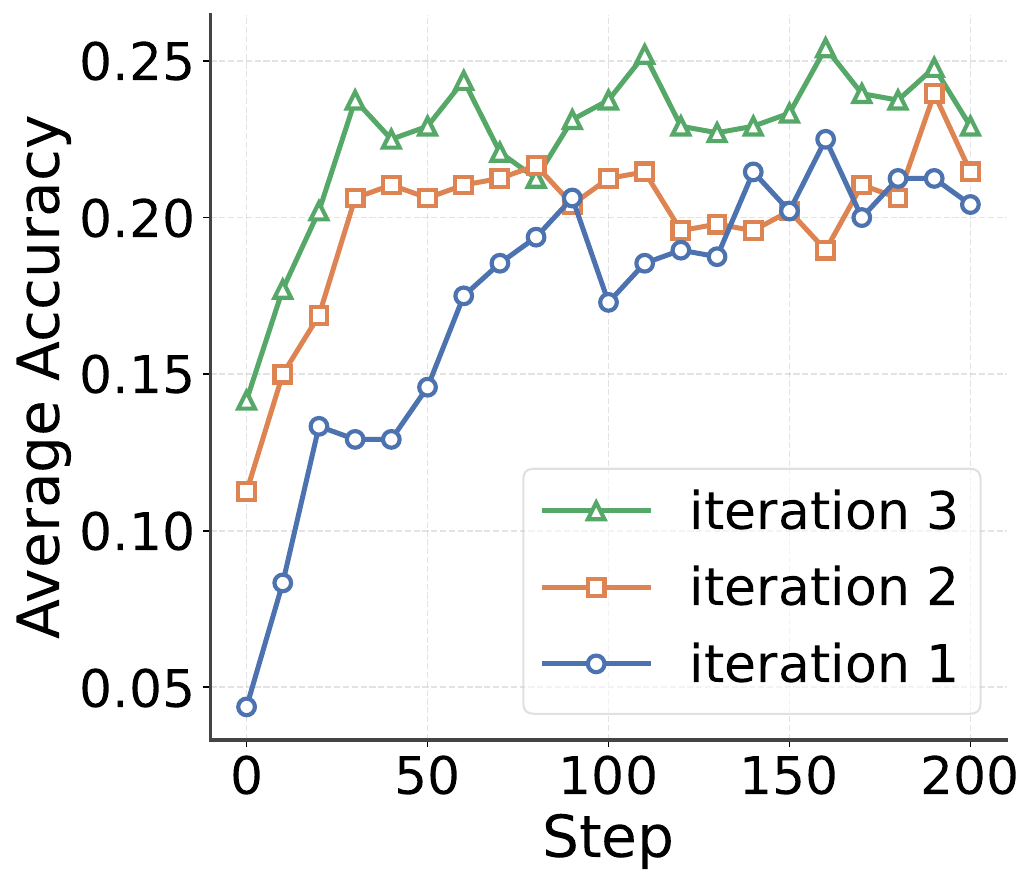}
    \caption{Qwen3-4B-Base}
    \label{fig:iter_4b}
  \end{subfigure}
  \caption{Accuracy curves of two models on AIME25 across training iterations.}
  \label{fig:iter_acc}
\end{figure}

We further examine the role of iterative training itself. As shown in Figure~\ref{fig:iter_acc}, from iteration 1 to 3, each successive round starts from a higher initial accuracy and maintains a stronger performance curve over most training steps. This suggests that experience accumulated in earlier iterations provides increasingly useful guidance for later training, highlighting the importance of iterative training in DGO. Appendix~\ref{Quality_Experiences} provides a case study showing how experience quality improves across iterations.


\section{Analysis}
In this section, we analyze the role of experience in our method from two complementary aspects: \emph{utilization} and \emph{internalization}. We also provide case studies in Appendix~\ref{uti_case} and Appendix~\ref{inte_case} to offer intuitive evidence for both capabilities.

\subsection{The Role of Experience in Utilization}
\label{utilization}


\begin{figure}[t]
  \centering
  \begin{subfigure}[b]{0.3\linewidth}
    \centering
    \includegraphics[width=\linewidth]{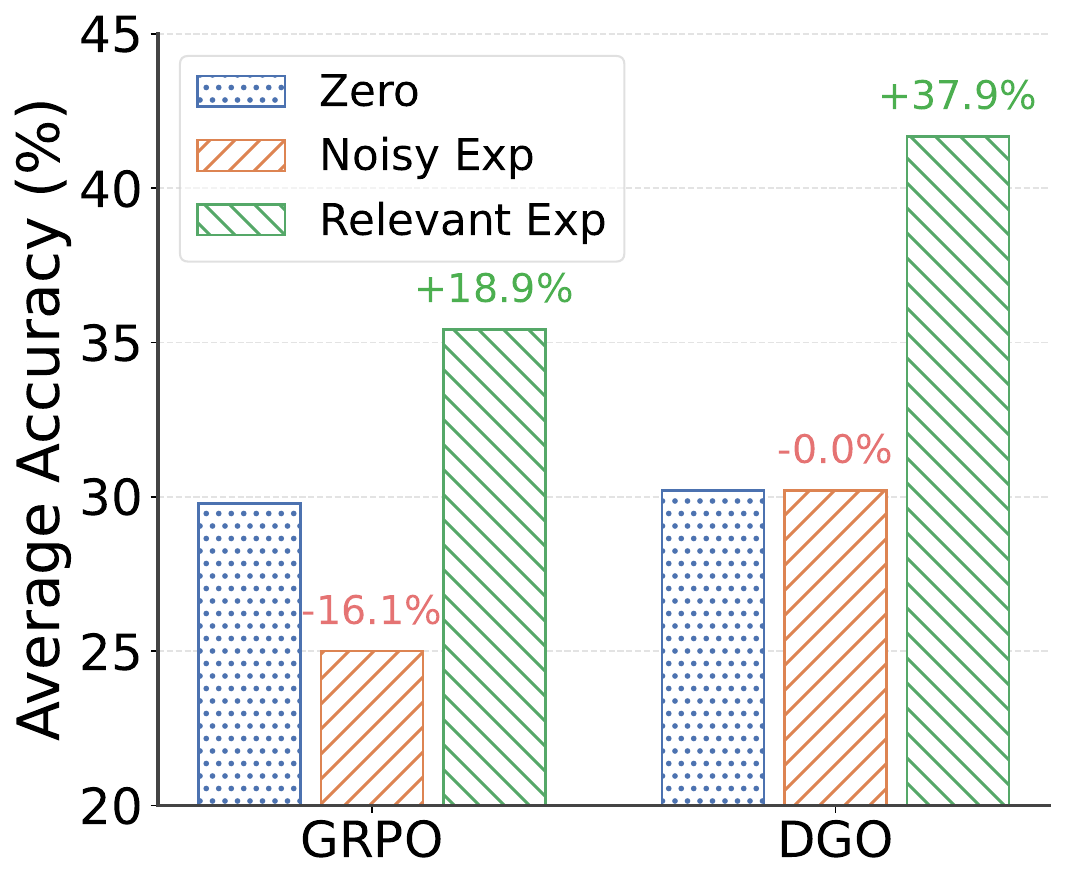}
    \caption{Qwen-8B-Base}
    \label{fig:passk}
  \end{subfigure}
  \hspace{0.03\linewidth}
  \begin{subfigure}[b]{0.3\linewidth}
    \centering
    \includegraphics[width=\linewidth]{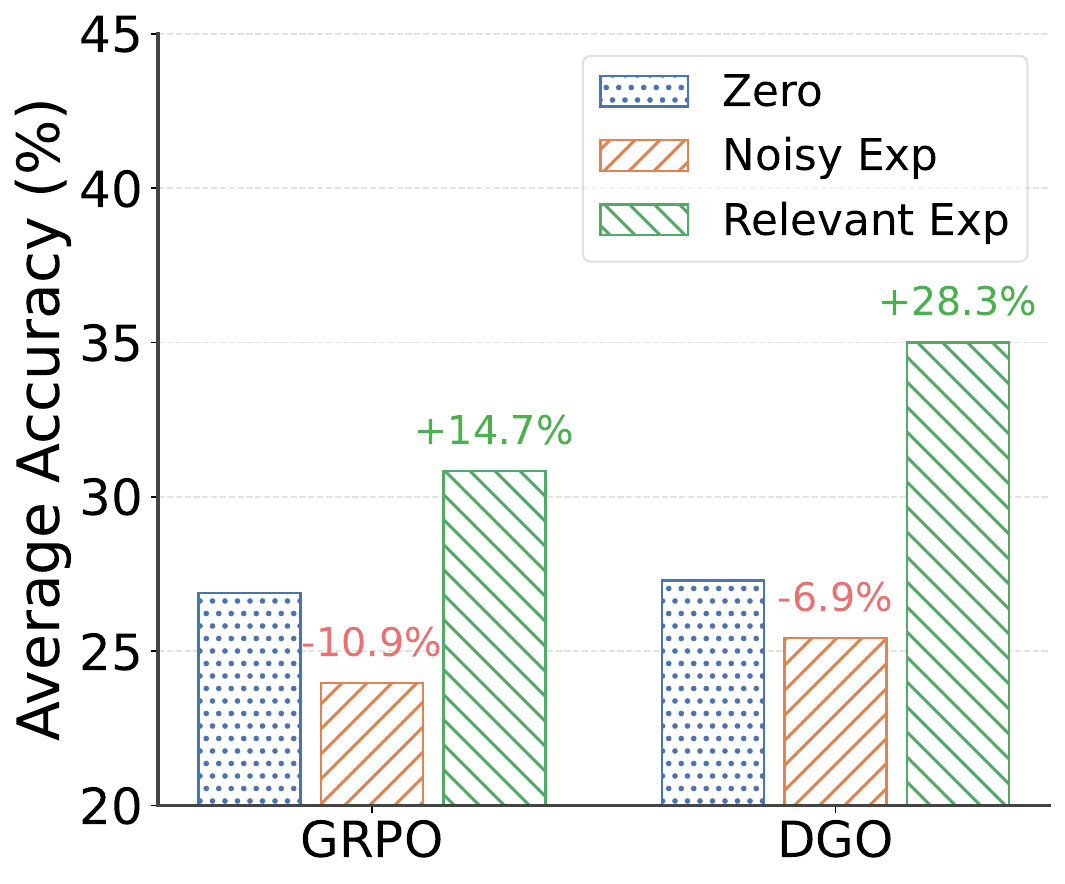}
    \caption{Qwen-4B-Base}
    \label{fig:avgk}
  \end{subfigure}
  \caption{Experience utilization under three settings on AIME24. DGO benefits more from relevant experience and is more robust to irrelevant experience. Numbers denote relative improvement over zero experience.}
  \label{fig:leverage_bar}
\end{figure}

Even high-quality experience cannot be fully exploited if the model cannot effectively utilize it~\cite{dou2026clbenchbenchmarkcontextlearning}. We compare DGO- and GRPO-trained models under three settings: \emph{without experience}, \emph{with noisy experience}, and \emph{with relevant experience}. Relevant experience is summarized from historical trajectories on the same problem, while noisy experience is from those of irrelevant problems. The same experience is provided to both models in each setting.

\paragraph{DGO makes more effective use of relevant experience.}
As shown in Figure~\ref{fig:leverage_bar}, adding relevant experience yields substantially larger gains for DGO. On Qwen-4B-Base, the improvement is 28.3\% for DGO, compared with 14.7\% for GRPO; on Qwen-8B-Base, the gain is 37.9\% for DGO versus 18.9\% for GRPO. This indicates that DGO is better at extracting useful signals from relevant experience.

\paragraph{DGO is more robust to noisy experience.}
Injecting noisy experience leads to a noticeably smaller degradation for DGO than for GRPO on both model scales, and on Qwen-8B-Base the performance of DGO is nearly unchanged. This suggests that DGO not only benefits more from helpful experience, but also resists distraction from irrelevant experience more effectively.

\subsection{The Role of Experience in Internalization}
\label{internalization}

Experience internalization refers to absorbing useful reasoning patterns into model parameters rather than relying on explicit experience at inference time. This effect is reflected in Table~\ref{tab:main_1}, where DGO consistently outperforms GRPO under intrinsic inference without external experience. Figure~\ref{fig:iter_acc} further shows that later training iterations start from higher initial accuracy and maintain better performance, suggesting that experience from earlier rounds is progressively retained by the model. Together, these results suggest that DGO improves not only experience utilization but also the internalization of reusable reasoning patterns.

\begin{figure}[t]
  \centering
  \begin{subfigure}[b]{0.3\linewidth}
    \centering
    \includegraphics[width=\linewidth]{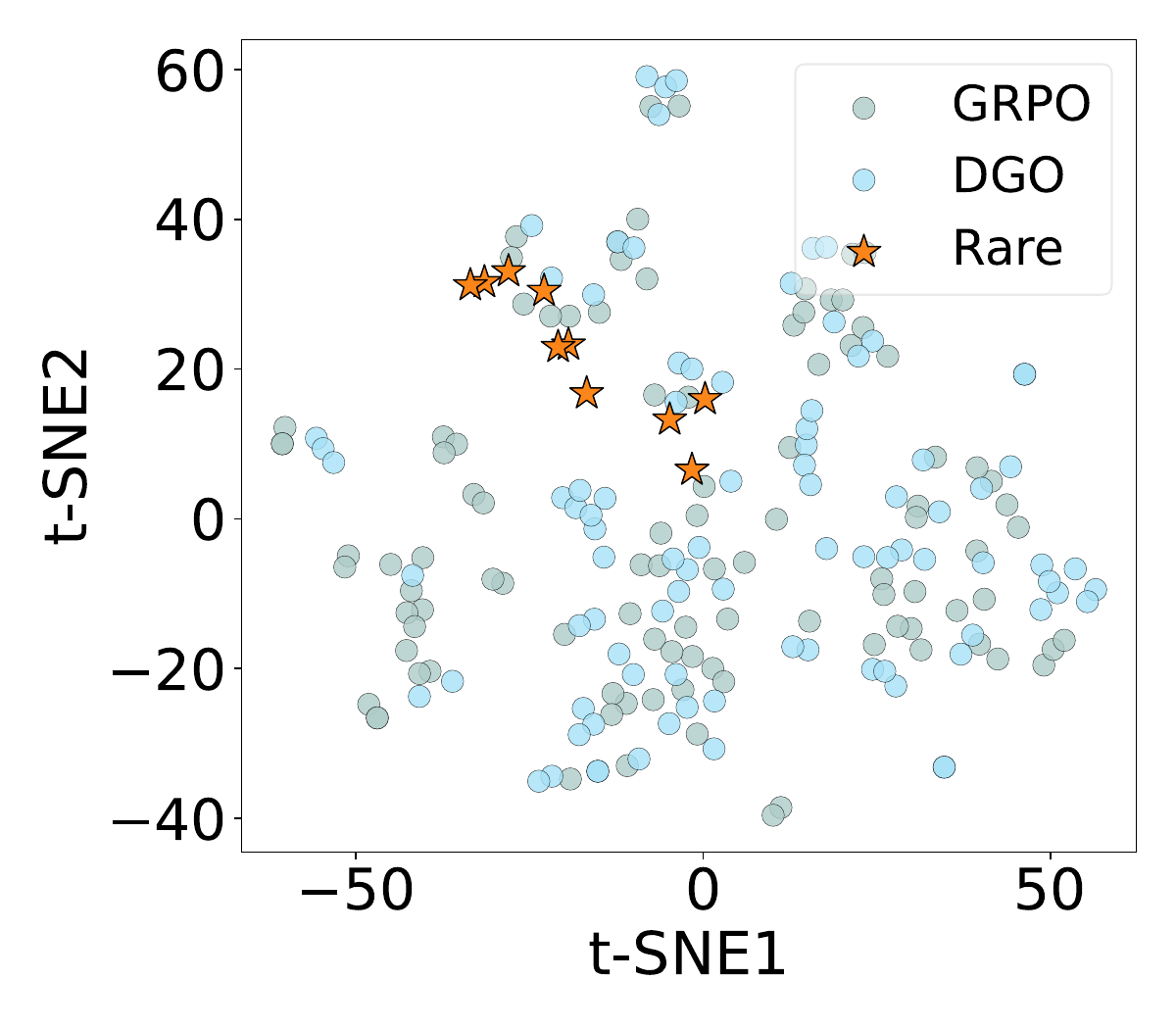}
    \caption{Qwen3-8B-Base}
    \label{fig:iter_8b}
  \end{subfigure}
  \hspace{0.03\linewidth}
  \begin{subfigure}[b]{0.3\linewidth}
    \centering
    \includegraphics[width=\linewidth]{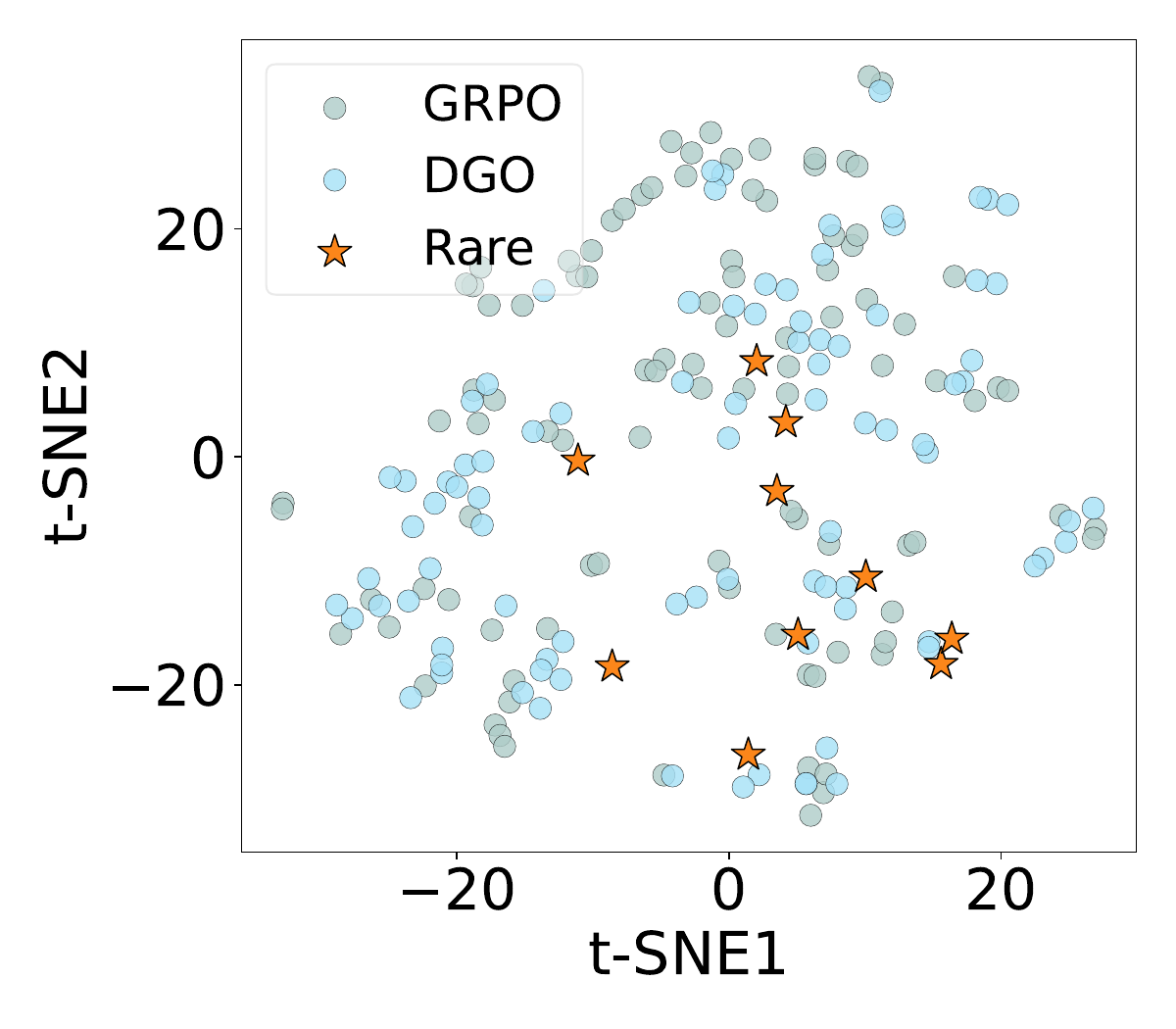}
    \caption{Qwen3-4B-Base}
    \label{fig:iter_4b}
  \end{subfigure}
  \caption{t-SNE visualization of generated trajectories on AIME24 for the DGO- and GRPO-trained models.}
  \label{fig:sne}
\end{figure}

\begin{figure}[t]
  \centering
  \begin{subfigure}[b]{0.3\linewidth}
    \centering
    \includegraphics[width=\linewidth]{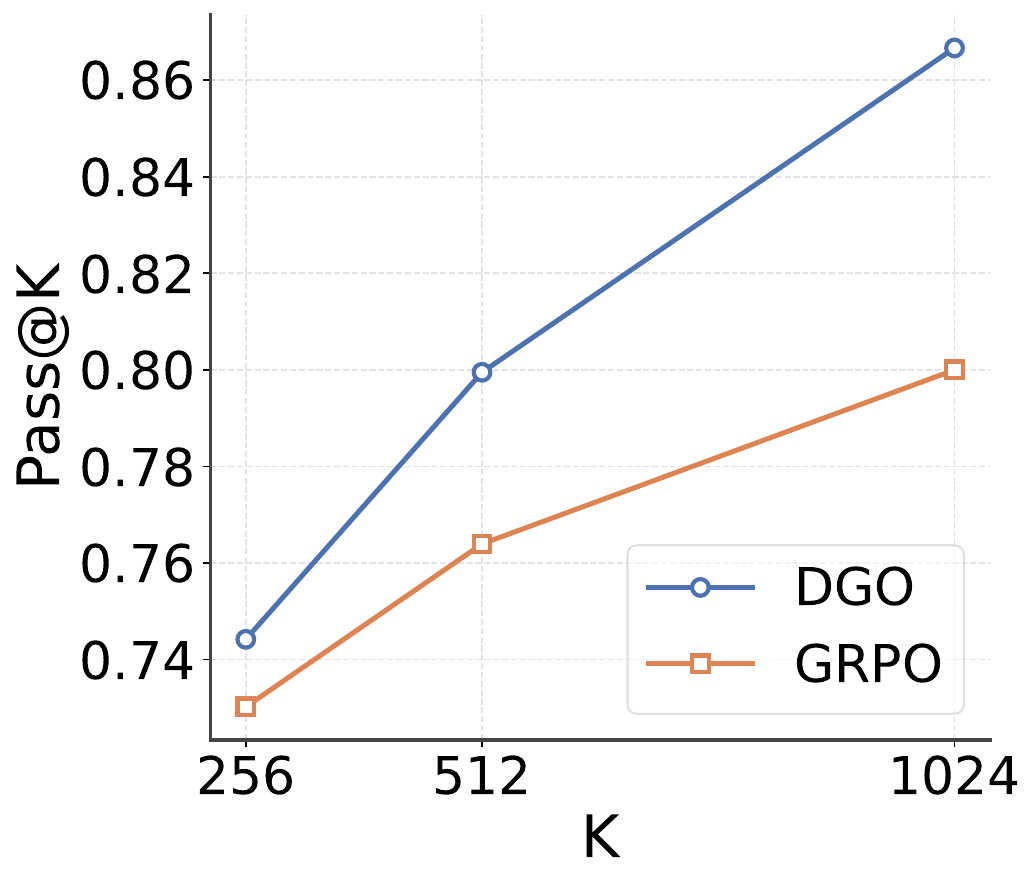}
    \caption{Qwen3-8B-Base}
    \label{fig:passk}
  \end{subfigure}
  \hspace{0.03\linewidth}
  \begin{subfigure}[b]{0.3\linewidth}
    \centering
    \includegraphics[width=\linewidth]{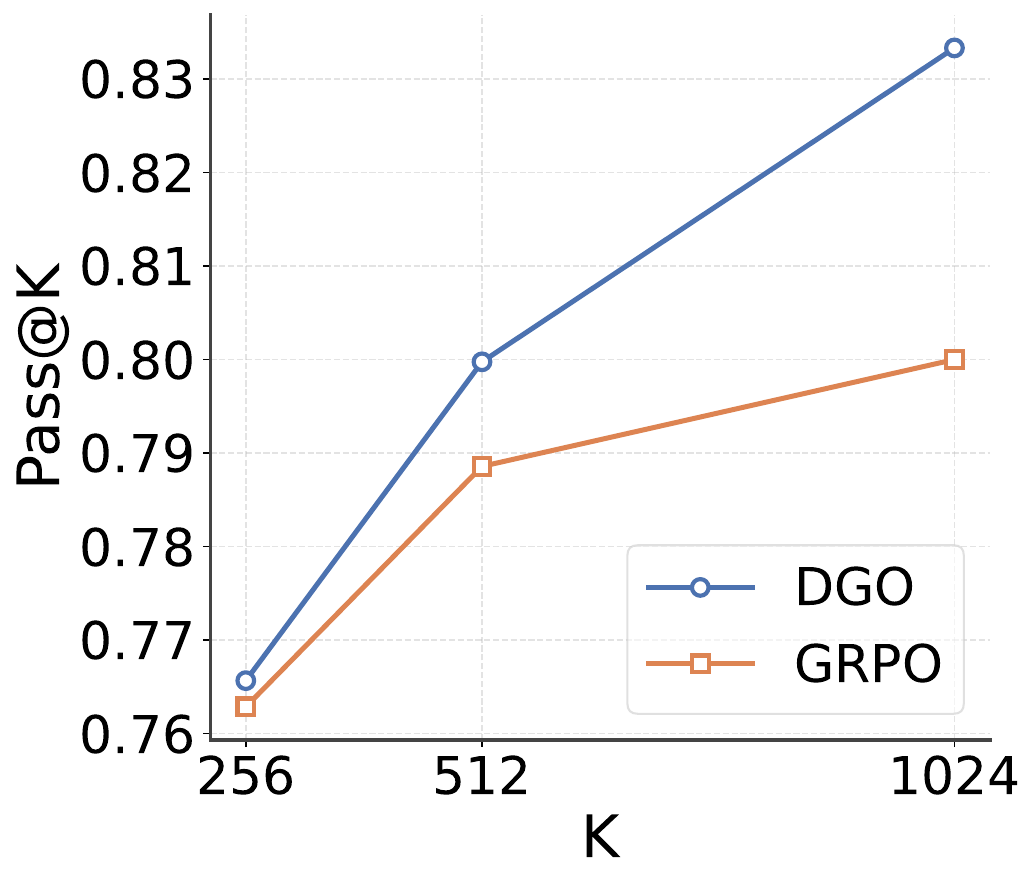}
    \caption{Qwen3-4B-Base}
    \label{fig:avgk}
  \end{subfigure}
  \caption{Comparison of the DGO- and GRPO-trained models on AIME24 in terms of Pass@K Accuracy.}
  \label{fig:passk_whole}
\end{figure}

To further examine what is internalized, we compare DGO and GRPO from two behavior-level perspectives. First, we analyze their trajectories in a semantic space. Following prior work~\cite{cheng2025reasoning}, \emph{rare trajectories} denote the top 10\% of DGO trajectories farthest from the GRPO distribution by average distance to their $k{=}5$ nearest neighbors. Second, we use Pass@K to test whether these behaviors correspond to valid solution modes reachable during sampling~\cite{yue2025does}.

\paragraph{DGO broadens the range of valid reasoning modes.}
As shown in Figure~\ref{fig:sne}, DGO produces trajectories that are semantically isolated from the GRPO distribution, suggesting reasoning behaviors beyond those induced by standard RL alone. Figure~\ref{fig:passk_whole} further shows that DGO consistently achieves higher Pass@K than GRPO across sampling budgets, indicating that these behaviors correspond to valid solution modes reachable during generation.
\section{Conclusion}

In this paper, we proposed \textbf{DGO}, a unified framework that improves LLM reasoning through dual guidance from external and internal experience. DGO integrates experience construction, joint trajectory-policy refinement, and experience internalization into an iterative process, enabling the model to better utilize external experience and progressively absorb useful reasoning patterns into its parameters. Experiments on multiple benchmarks show that DGO consistently improves both intrinsic reasoning and test-time scaling performance.
Future work includes enabling models to manage and update their own experience more autonomously, and to coordinate external and internal experience more effectively. This may further improve the efficiency, robustness, and sustainability of experiential learning in LLMs. Project Link: \url{https://github.com/RUCAIBox/DualGuidanceOptimization}.


\bibliography{ref}

\appendix

\section{Training Details}
\label{training_details}
\subsection{RL Stage}
To ensure stability and effectiveness in our RL experiments, we adopt the following configuration. Please refer to Table~\ref{tab:rl hyperparams}. All training experiments are implemented on 32 $\times$ H20 GPUs.


\begin{table}[t]
\centering
\small
\renewcommand{\arraystretch}{1.15}
\setlength{\tabcolsep}{12pt}
\begin{tabular}{l c}
\toprule
\textbf{Hyper-parameter} & \textbf{Value} \\
\midrule
Learning Rate            & 1e-6  \\
Total Steps              & 200   \\
Batch Size               & 128   \\
Mini Batch Size          & 32    \\
KL Loss Coefficient      & 0.0 \\
Clip Higher               & (0.28, 0.2)   \\
Temperature              & 1.0     \\
Number of Rollouts       & 16     \\
Maximum Prompt Length    & 2K  \\
Maximum Response Length  & 12K \\
\bottomrule
\end{tabular}
\caption{RL Training hyper-parameters.}
\label{tab:rl hyperparams}
\end{table}

\subsection{Internalization Stage}

\begin{table}[t]
\centering
\small
\renewcommand{\arraystretch}{1.15}
\setlength{\tabcolsep}{12pt}
\begin{tabular}{l c}
\toprule
\textbf{Hyper-parameter} & \textbf{Value} \\
\midrule
Learning Rate            & 1e-6  \\
Total Epochs              & 4   \\
Batch Size               & 128 for 8B/14B; 64 for 4B    \\
Maximum Length  & 6144 \\
\bottomrule
\end{tabular}
\caption{SFT Training hyper-parameters.}
\label{tab:sft hyperparams}
\end{table}

We show the internalization training configurations in Table~\ref{tab:sft hyperparams}.

\subsection{Experience Generator Training}
\begin{table}[t]
\centering
\small
\renewcommand{\arraystretch}{1.15}
\setlength{\tabcolsep}{12pt}
\begin{tabular}{l c}
\toprule
\textbf{Hyper-parameter} & \textbf{Value} \\
\midrule
Learning Rate            & 1e-5  \\
Total Epochs              & 4   \\
Batch Size               & 128    \\
Maximum Length  & 16384 \\
\bottomrule
\end{tabular}
\caption{Experience Generator Training hyper-parameters.}
\label{tab:exp gen hyperparams}
\end{table}

We show the training configurations of experience generator in Table~\ref{tab:exp gen hyperparams}. The training data is sourced from SkyWork as well.

\section{Prompt Template}
\label{app:prompt}

\subsection{Prompt for Experience Extraction}
\begin{table*}[t]
\centering
\begin{tabular}{p{0.95\textwidth}}
\toprule
\begin{minipage}{0.93\textwidth}
\begin{Verbatim}[breaklines=true, breakanywhere=true, fontsize=\scriptsize]
### You are an expert knowledge extractor specializing in converting problem-solving trajectories into abstract, reusable reasoning rules that preserve crucial mathematical specificity and improve future problem solving efficiency.

Your task is to analyze a given problem and a candidate solution trajectory. Based on logical consistency, mathematical validity, and alignment with the problem requirements, judge whether the solution is likely correct or incorrect, and then extract **one concise general rule** as reusable experience.

The solution may be correct or incorrect. You must infer this by critically examining the reasoning and results.

#### Instructions:

1. **Input Analysis**:
    **problem**: The original problem statement.
    **solution**: A full solution trajectory attempting to solve the problem.

2. **Experience Extraction Guidelines**:
    - **Self-judgment**:
        Evaluate whether the solution is likely correct by checking:
        (a) whether the final answer satisfies the problem conditions, and  
        (b) whether each reasoning step is logically and mathematically valid.

    - **If the solution appears correct**:
        Do NOT merely restate why it works.
        Instead, focus on extracting **optimization experience**, such as:
            • how the solution could be made shorter or more direct,  
            • which intermediate steps, case splits, or calculations are redundant,  
            • which key insight could be used earlier to avoid detours,  
            • how to reformulate the approach into a cleaner standard method.
        Summarize this as a rule that helps future solvers reach the answer **faster and with less complexity**.

    - **If the solution appears incorrect**:
        Identify the most critical error, unjustified assumption, or logical gap.
        Extract an **avoiding-errors experience** that explains:
            • what warning sign to look for, and  
            • what should be done instead to prevent this type of mistake.

    - **One Output**:
        Generate **exactly one rule** that reflects either:
        (a) an optimization rule distilled from a correct but improvable solution, or  
        (b) an avoiding-errors rule distilled from an incorrect solution.

    - **Abstraction with Specificity**:
        Generalize the rule while retaining essential mathematical structures, formulas, constraints, or thresholds that make the experience precise and actionable.
        Use LaTeX for mathematical expressions where appropriate.

3. **Rule Format Requirement**:
    Output **exactly one rule** in the following structure, with no extra text:
        WHEN <Broad domain or phase>.
        IF <Specific condition, scenario, or inefficiency / potential error>.
        THEN <Action to take or avoid, emphasizing simplification, direct strategy, or error prevention, with key formulas if needed>.

# **Input:**
# problem:\n{problem}\n
# solution:\n{solution}\n

# **Output:**
WHEN <Broad domain or phase>. IF <Specific condition, scenario, or inefficiency / potential error>. THEN <Action to take or avoid>.
\end{Verbatim}
\end{minipage} \\
\bottomrule
\end{tabular}
\caption{Prompt template for experience extraction.}
\label{tab:experience_prompt}
\end{table*}

Our prompt for experience extraction is shown in Table~\ref{tab:experience_prompt}.

\subsection{Prompt for Reasoning}
\subsubsection{Reasoning without Experience}
\begin{table*}[t]
\centering
\begin{tabular}{p{0.95\textwidth}}
\toprule
\begin{minipage}{0.93\textwidth}
\begin{Verbatim}[breaklines=true, breakanywhere=true, fontsize=\scriptsize]
Please reason step by step, and put your final answer within \boxed{{}}.
{problem}
\end{Verbatim}
\end{minipage} \\
\bottomrule
\end{tabular}
\caption{Prompt template for reasoning without experience.}
\label{tab:reason_wo_exp}
\end{table*}

Our prompt for reasoning without experience is shown in Table~\ref{tab:reason_wo_exp}.

\subsubsection{Reasoning with Experience}
\begin{table*}[t]
\centering
\begin{tabular}{p{0.95\textwidth}}
\toprule
\begin{minipage}{0.93\textwidth}
\begin{Verbatim}[breaklines=true, breakanywhere=true, fontsize=\scriptsize]
Please reason step by step, and put your final answer within \boxed{{}}. You will be given some reusable problem-solving experiences, and you can refer to them during the reasoning.

[Experiences]
{experiences}

[Problem]
{problem}
\end{Verbatim}
\end{minipage} \\
\bottomrule
\end{tabular}
\caption{Prompt template for reasoning with experience.}
\label{tab:reason_w_exp}
\end{table*}

Our prompt for reasoning with experience is shown in Table~\ref{tab:reason_w_exp}.

\subsection{Prompt for Rephrasing}

\begin{table*}[t]
\centering
\begin{tabular}{p{0.95\textwidth}}
\toprule
\begin{minipage}{0.93\textwidth}
\begin{Verbatim}[breaklines=true, breakanywhere=true, fontsize=\scriptsize]
You are an expert mathematician. You are given a math problem, some background [Experience Content], and a [Reference Solution] which may have used those experiences explicitly.

Your task is to re-author a "Gold Standard" solution. This solution must be entirely self-contained, as if you solved the problem using only your innate mathematical intuition.

Strict requirements:

1. Total De-Reference: You MUST NOT mention, quote, or even subtly refer to the [Experience Content] provided. Do not use phrases like "By logic of...", "Following the principle of...", or any terminology that implies you are looking at an external guide. The reader should have no idea that any "Experience Content" was provided to you.

2. Invisible Integration: Instead of "using" the experience, you must "be" the expert who already knows it. If the experience suggests a shortcut or a formula, justify its use through first-principles reasoning within the flow of the solution. For example, do not say "Using the formula for X," say "Since the problem requires [reason], we can derive that [logical step]..."

3. Silent Audit & Correction: Treat the [Reference Solution] as a potentially flawed scratchpad. Verify every calculation and logical leap. If it is wrong, fix it silently. Your final output must be a clean, error-free, and authoritative proof. Do NOT mention that the reference was wrong.

4. Strategic Fidelity: Follow the general strategy of the [Reference Solution] (e.g., if it uses energy conservation, you should too), but execute it with perfect precision and without any meta-talk or conversational fillers.

5. Final Answer Consistency: The final numerical/symbolic answer must exactly match the [Reference Solution] and be enclosed in \boxed{{}}.

[Experience Content]
{experiences}

[Problem]
{problem}

[Reference Solution]
{solution}

[Your Solution]
\end{Verbatim}
\end{minipage} \\
\bottomrule
\end{tabular}
\caption{Prompt template for solution rewriting.}
\label{tab:rewrite_prompt}
\end{table*}

Our prompt for solution rephrasing is shown in Table~\ref{tab:rewrite_prompt}.

\section{Case Study}
\subsection{Example of Experience}
\begin{table*}[t]
\centering
\begin{tabular}{p{0.95\textwidth}}
\toprule
\begin{minipage}{0.93\textwidth}
\begin{Verbatim}[breaklines=true, breakanywhere=true, fontsize=\scriptsize]
{
  "problem": "Let \(b\ge 2\) be an integer. Call a positive integer \(n\) \(b\text-\textit{eautiful}\) if it has exactly two digits when expressed in base \(b\) and these two digits sum to \(\sqrt n\). For example, \(81\) is \(13\text-\textit{eautiful}\) because \(81 = \underline{6}\ \underline{3}_{13}\) and \(6 + 3 = \sqrt{81}\). Find the least integer \(b\ge 2\) for which there are more than ten \(b\text-\textit{eautiful}\) integers.",
  "experiences": [
    "WHEN solving problems about numbers with given digit sums and square roots in a base representation.\nIF setting up an equation from the condition that the sum of digits equals the square root of the number,\nTHEN ensure the correct conversion from base \(b\) to decimal: if the two-digit representation is \(\overline{ad}_b\) with digits \(a\) (tens) and \(d\) (units), then \(n = ab + d\), and the condition is \(a + d = \sqrt{ab + d}\). Square both sides to get \((a+d)^2 = ab + d\), then rearrange to analyze integer solutions, noting digit constraints \(1 \le a \le b-1\), \(0 \le d \le b-1\). Avoid incorrectly writing \(n = b(a+1)\) which omits the units digit.",
    "WHEN solving for numbers with digit sum equal to the square root in a given base.\nIF the condition leads to \(n = k^2\) and \(d_1 = \frac{k(k-1)}{b-1}\) where \(d_1\) is the leading digit in base \(b\).\nTHEN use the divisibility condition \((b-1) \mid k(k-1)\) to find all \(k\) for each \(b\) by factoring \(b-1\), which is more efficient than brute-forcing over the range of \(k\).",
    "WHEN solving for the smallest base \( b \) with a specified count of numbers satisfying a digit-sum and square root condition.\nIF you have derived the relation \( d_1 = \frac{S(S-1)}{b-1} \) with \( S = d_1 + d_2 \), and \( b-1 \le S^2 < b^2 \).\nTHEN iterate over integer \( S \ge 2 \), for each \( S \) compute all divisors \( d \) of \( S(S-1) \), set \( b = d+1 \), and check if \( b-1 \le S^2 < b^2 \); if so, record \( (b,S) \). Then collect all \( b \) and their counts to efficiently find the smallest \( b \) with the required count.",
    "WHEN solving problems about numbers with digit conditions in a given base.\nIF the condition leads to an equation of the form $d_1 = \frac{k^2 - k}{b-1}$ with $k = d_1 + d_2$ and $d_1$ must be an integer between $1$ and $b-1$,\nTHEN systematically check all integers $k$ from $\lceil \sqrt{b-1} \rceil$ to $b-1$, and verify that $(b-1) \mid k(k-1)$; this ensures a complete count without missing valid solutions."
  ]
}
\end{Verbatim}
\end{minipage} \\
\bottomrule
\end{tabular}
\caption{An example of retrieved experiences associated with a problem.}
\label{tab:experience_example}
\end{table*}

We present several examples of experiences generated by our experience generator in Table~\ref{tab:experience_example}.

\subsection{Comparison of Utilization between DGO and GRPO}
\label{uti_case}
\begin{table*}[t]
\centering
\begin{tabular}{p{0.95\textwidth}}
\toprule
\begin{minipage}{0.93\textwidth}
\begin{Verbatim}[breaklines=true, breakanywhere=true, fontsize=\scriptsize]
[Problem]
Let \omega\neq 1 be a 13th root of unity. Find the remainder when \[ \prod_{k=0}^{12}(2-2\omega^k+\omega^{2k}) \] is divided by 1000.

[Experiences]
[1] WHEN solving set-counting problems with multiple sets and given counts of exactly k-set intersections.
IF the total number of elements and the sizes of each set are known, and the goal is to find the intersection of all sets.
THEN compute the total number of set memberships \( T = \sum |A_i| \) and equate it to \( \sum_k k \cdot n_k \), where \( n_k \) is the number of elements in exactly \( k \) sets, leading to a linear equation for the unknown intersection count.


\end{Verbatim}
\end{minipage} \\
\bottomrule
\end{tabular}

\vspace{0.4em}

\begin{tabular}{p{0.465\textwidth} p{0.465\textwidth}}
\toprule
\textbf{DGO-trained Trajectory~(Correct)} & \textbf{GRPO-trained Trajectory~(Incorrect)} \\
\midrule

\begin{minipage}[t]{0.44\textwidth}
\begin{Verbatim}[breaklines=true, breakanywhere=true, fontsize=\scriptsize]
When \(k=0\), the factor is
\[
2-2(1)+1^2=1,
\]
so it remains to evaluate
\[
\prod_{k=1}^{12}(2-2\omega^k+\omega^{2k}).
\]

Let
\[
P(x)=x^2-2x+2=(x-(1+i))(x-(1-i)).
\]

Then
\[
\prod_{k=1}^{12}P(\omega^k)
=
\left(\prod_{k=1}^{12}(\omega^k-(1+i))\right)
\left(\prod_{k=1}^{12}(\omega^k-(1-i))\right).
\]

Using
\[
\prod_{y^{13}=1,\ y\neq 1}(y-a)=\frac{a^{13}-1}{a-1},
\]
the product becomes
\[
[(1+i)^{13}-1][(1-i)^{13}-1].
\]

(-65-64i)(-65+64i)=8321.
\]

Hence,
\[
8321 \bmod 1000 = 321.
\]

\[
\boxed{321}
\]
\end{Verbatim}
\end{minipage}
&
\begin{minipage}[t]{0.44\textwidth}
\begin{Verbatim}[breaklines=true, breakanywhere=true, fontsize=\scriptsize]
Let
\[
f(y)=2-2y+y^2.
\]

We need
\[
P=\prod_{y\in\mu_{13}\setminus\{1\}} (2-2y+y^2).
\]

Consider
\[
Q(x)=\prod_{y\in\mu_{13}\setminus\{1\}}(x-y)
=\frac{x^{13}-1}{x-1}.
\]

Now, we can express \(P\) in terms of \(Q(x)\). Notice that
\[
P=Q(2).
\]

Thus
\[
P=2^{12}+2^{11}+\cdots+2+1=8191,
\]
so the remainder is
\[
8191 \bmod 1000=191.
\]

The trajectory later rewrites
\[
2-2y+y^2=(y-1)^2+1,
\]
but does not derive a valid product identity from it, and
still returns
\[
\boxed{191}.
\]
\end{Verbatim}
\end{minipage}
\\
\bottomrule
\end{tabular}

\caption{Comparison of two trajectory excerpts under the same retrieved experiences. Both trajectories receive the same irrelevant experiences, but the left one stays problem-driven while the right one makes an invalid polynomial substitution.}
\label{tab:case_noise_compare}
\end{table*}

In Table~\ref{tab:case_noise_compare}, both the DGO-trained model and the GRPO-trained model are provided with the same retrieved experiences, all of which are clearly irrelevant to the current problem. The experiences concern set counting and inclusion-exclusion, whereas the target problem is a product over roots of unity. This setup therefore serves as a direct test of robustness to noisy retrieved experience.

The DGO-trained model remains largely problem-driven despite the presence of irrelevant experience. Its trajectory quickly identifies the algebraic structure of the expression, factorizes \(x^2-2x+2\) into \((x-(1+i))(x-(1-i))\), and then applies the standard roots-of-unity product identity to transform the original product into evaluations at \(1\pm i\). The overall reasoning path is coherent and tightly aligned with the mathematical structure of the problem. Importantly, although noisy experiences are present in the prompt, they do not meaningfully alter the solution strategy. This suggests that the DGO-trained model is able to filter out irrelevant external guidance and rely on intrinsically relevant structure when the retrieved experience is unhelpful.

By contrast, the GRPO-trained model is more vulnerable to the same noisy experience. Although it begins with a superficially similar setup, it fails to maintain a stable problem-specific line of reasoning and eventually makes an invalid substitution, incorrectly reducing the target product to \(Q(2)\). This error reflects not simply a calculation mistake, but a weaker ability to discriminate between structurally relevant and irrelevant guidance. Under noisy retrieval, the model appears more prone to drifting toward a loose template-based manipulation rather than preserving the exact algebraic requirements of the problem. As a result, it reaches the wrong answer.

This comparison highlights an important difference in experience utilization. Stronger utilization is not merely the ability to use retrieved experience when it is helpful, but also the ability to reject or suppress it when it is irrelevant. The DGO-trained model demonstrates better robustness under noisy experience, whereas the GRPO-trained model is more easily distracted by irrelevant context. This suggests that DGO improves not only the effectiveness of leveraging useful experience, but also the selectivity required to avoid being misled by unhelpful experience.

\subsection{Comparison of Internalization between DGO and GRPO}
\label{inte_case}
\begin{table*}[t]
\centering
\begin{tabular}{p{0.95\textwidth}}
\toprule
\begin{minipage}{0.93\textwidth}
\begin{Verbatim}[breaklines=true, breakanywhere=true, fontsize=\scriptsize]
[Problem]
There exist real numbers \(x\) and \(y\), both greater than 1, such that
\[
\log_x\left(y^x\right)=\log_y\left(x^{4y}\right)=10.
\]
Find \(xy\).
\end{Verbatim}
\end{minipage} \\
\bottomrule
\end{tabular}

\vspace{0.4em}

\begin{tabular}{p{0.465\textwidth} p{0.465\textwidth}}
\toprule
\textbf{DGO-trained Trajectory~(Correct)} & \textbf{GRPO-trained Trajectory~(Incorrect)} \\
\midrule

\begin{minipage}[t]{0.44\textwidth}
\begin{Verbatim}[breaklines=true, breakanywhere=true, fontsize=\scriptsize]
From
\[
\log_x(y^x)=10,
\]
we get
\[
x\log_x y=10.
\]

Let
\[
a=\log_x y.
\]
Then
\[
x=\frac{10}{a}.
\]

From
\[
\log_y(x^{4y})=10,
\]
we get
\[
4y\log_y x=10.
\]

Let
\[
b=\log_y x.
\]
Then
\[
y=\frac{5}{2b}.
\]

Using the reciprocal relation
\[
\log_y x=\frac{1}{\log_x y},
\]
we have
\[
b=\frac{1}{a}.
\]

Thus
\[
y=\frac{5}{2\cdot (1/a)}=\frac{5a}{2}.
\]

Therefore,
\[
xy=\frac{10}{a}\cdot \frac{5a}{2}=25.
\]

\[
\boxed{25}
\]
\end{Verbatim}
\end{minipage}
&
\begin{minipage}[t]{0.44\textwidth}
\begin{Verbatim}[breaklines=true, breakanywhere=true, fontsize=\scriptsize]
From
\[
\log_x(y^x)=10,
\]
the solution rewrites
\[
y=x^{10/x}.
\]

From
\[
\log_y(x^{4y})=10,
\]
it rewrites
\[
x=y^{5/(2y)}.
\]

Then it substitutes one equation into the other:
\[
x=\left(x^{10/x}\right)^{5/(2x^{10/x})}
  =x^{25/(x^{10/x+1})},
\]
and concludes
\[
x^{10/x+1}=25.
\]

Instead of using the reciprocal relation between
\(\log_x y\) and \(\log_y x\), the trajectory then turns to
repeated numerical guessing and fails to obtain a stable
symbolic simplification.

The repeated trial steps are omitted here.

Final answer:
\[
\boxed{\frac{3}{10}}
\]
\end{Verbatim}
\end{minipage}
\\
\bottomrule
\end{tabular}

\caption{Comparison of two trajectory excerpts without retrieved experience. The DGO-trained model directly invokes a compact logarithmic relation and reaches the correct result, while the GRPO-trained model fails to internalize the same simplification pattern and falls into unstable symbolic manipulation.}
\label{tab:case_internalization_compare}
\end{table*}

In Table~\ref{tab:case_internalization_compare}, no external experience is provided at inference time, so the difference between the two trajectories mainly reflects the extent to which useful reasoning patterns have been internalized into the model parameters. The problem itself is structurally simple: after expanding the logarithms, the key step is to introduce \(\log_x y\) and \(\log_y x\) and then use the reciprocal relation between them. Once this relation is recognized, the quantity \(xy\) collapses immediately to a constant.

The DGO-trained model exhibits exactly this kind of internalized competence. Its trajectory quickly abstracts the problem into two compact symbolic relations, explicitly introduces auxiliary variables, and then invokes the identity \(\log_y x = 1/\log_x y\) to connect the two equations. The subsequent derivation is short, stable, and fully symbolic, leading directly to \(xy=25\). Notably, the model does not rely on trial-and-error search or unnecessary algebraic expansion. This suggests that the relevant reasoning pattern has been internalized as a reusable parametric capability.

By contrast, the GRPO-trained model does not show the same level of internalization. Although it begins from the same logarithmic expansions, it fails to consolidate them into the reciprocal-log structure that makes the problem easy. Instead, it substitutes one equation into the other, produces a complicated self-referential expression, and then falls back to repeated numerical guessing. The trajectory becomes increasingly unstable and eventually loses focus altogether. This behavior indicates that the model has not fully absorbed the underlying simplification pattern into its parameters, even though the problem can in principle be solved by a short symbolic argument.

This comparison reveals the role of experience internalization. Strong internalization is reflected not only in getting the correct answer, but in whether the model can spontaneously recover the right abstraction without external guidance. The DGO-trained model demonstrates stronger internalization by directly activating a compact and transferable reasoning schema, whereas the GRPO-trained model is more likely to remain trapped in local manipulations without reaching the core structural insight.

\subsection{Quality of Experience across Iterations}
\label{Quality_Experiences}
\begin{table*}[t]
\centering
\begin{tabular}{p{0.95\textwidth}}
\toprule
\begin{minipage}{0.93\textwidth}
\begin{Verbatim}[breaklines=true, breakanywhere=true, fontsize=\scriptsize]
[Problem]
Let \(b\ge 2\) be an integer. Call a positive integer \(n\) \(b\text-\textit{eautiful}\) if it has exactly two digits when expressed in base \(b\) and these two digits sum to \(\sqrt n\). For example, \(81\) is \(13\text-\textit{eautiful}\) because \(81 = \underline{6}\ \underline{3}_{13}\) and \(6 + 3 = \sqrt{81}\). Find the least integer \(b\ge 2\) for which there are more than ten \(b\text-\textit{eautiful}\) integers.

[Round 0]
WHEN solving for numbers with digit sum equal to the square root in a given base.
IF the condition leads to \(n = k^2\) and \(d_1 = \frac{k(k-1)}{b-1}\) where \(d_1\) is the leading digit in base \(b\).
THEN use the divisibility condition \((b-1) \mid k(k-1)\) to find all \(k\) for each \(b\) by factoring \(b-1\), which is more efficient than brute-forcing over the range of \(k\).

[Round 1]
WHEN solving problems about digit conditions in a given base.
IF the condition is \( a+d = \sqrt{ab+d} \) with \( a \) the leading digit, \( 1 \le a \le b-1 \), \( 0 \le d \le b-1 \).
THEN set \( k = a+d \) so \( n = k^2 \), derive \( a = \frac{k(k-1)}{b-1} \), and find \( b \) such that \( b-1 \mid k(k-1) \) for multiple \( k \), ensuring \( a \) and \( d = k-a \) satisfy digit bounds, to avoid inefficient brute-force enumeration.

[Round 2]
WHEN solving problems about numbers with two digits in base \(b\) satisfying a digit-sum condition involving \(\sqrt{n}\).
IF you have derived that \(n = a b + d\) with \(a+d = k = \sqrt{n}\) and obtained the relation \(a = \frac{k^2 - k}{b-1}\).
THEN note that \(k(k-1)\) must be divisible by \(b-1\), which is equivalent to \(k \equiv 0\) or \(1 \pmod{b-1}\); use this congruence to restrict \(k\) to two arithmetic progressions within the range \(\lceil \sqrt{b} \rceil \le k \le b-1\), drastically reducing the number of candidates to check.
\end{Verbatim}
\end{minipage} \\
\bottomrule
\end{tabular}
\caption{The problem and one representative extracted experience from each round.}
\label{tab:exp_evolution_example}
\end{table*}

As shown in Table~\ref{tab:exp_evolution_example}, across the three rounds, the extracted experience becomes progressively more structured, more actionable, and better aligned with the true computational bottleneck of the problem.

The Round 0 experience already captures the key algebraic reduction: the problem can be transformed into the divisibility condition \((b-1)\mid k(k-1)\) together with the formula \(d_1=\frac{k(k-1)}{b-1}\). This is a meaningful step beyond naive brute-force search, since it identifies the central arithmetic constraint underlying valid \(b\)-beautiful numbers. However, the experience is still relatively coarse-grained. It points to factoring \(b-1\) and using divisibility, but does not yet describe a complete reasoning chain from the original digit condition to a systematic counting procedure.

The Round 1 experience is more complete and operational. Instead of starting from an already reduced form, it begins from the original condition \(a+d=\sqrt{ab+d}\), explicitly introduces the substitution \(k=a+d\), derives \(a=\frac{k(k-1)}{b-1}\), and then reminds the solver to verify the digit constraints through \(d=k-a\). Compared with Round 0, this version better preserves the full reasoning pipeline from the original statement to the final verification conditions. As a result, it is more directly reusable as a solution template rather than merely a high-level hint.

The Round 2 experience goes one step further by making the counting structure itself more explicit. Rather than stopping at the divisibility condition, it turns that condition into a sharper congruence-based restriction on \(k\), and uses it to narrow the candidate space to a much smaller set. This makes the experience more closely matched to the real objective of the problem, which is not just to verify a single candidate, but to identify the smallest base \(b\) for which the number of valid solutions exceeds a threshold. In this sense, the third-round experience is the most strategic: it abstracts away from local derivation details and focuses on how to reduce the search space efficiently.

Overall, the progression from Round 0 to Round 2 shows a clear refinement trend. The experience evolves from identifying a useful arithmetic condition, to providing a full derivation-and-checking template, and finally to exposing the higher-level structural principle that supports efficient counting. This illustrates how iterative refinement can gradually transform raw problem-solving traces into more compact, transferable, and strategically valuable experience.

\section{Additional Details}
\subsection{Rule of Experience Update}
\label{Exp Update}
Experience update follows a redundancy-first, quality-second procedure:
\begin{itemize}
    \item \textbf{Redundancy check.} For each new experience, we first compute its semantic similarity with the existing experiences in the bank. If the maximum similarity is below a threshold $\beta$, the new experience is treated as a candidate addition. Otherwise, it is regarded as redundant with the most similar existing experience and is treated as a candidate replacement. We use a predefined similarity threshold $\beta=0.8$ for redundancy detection.
    \item \textbf{Quality check.} We then evaluate the candidate action on the validation set using avg@16. If the new experience is a candidate addition, it is added only when it improves the validation performance of the current bank. If it is a candidate replacement, it replaces the old experience only when it yields a higher avg@16 than the old one.
\end{itemize}

\subsection{Rule of Trajectory Filter}
\label{app:filter}
When rewriting experience-guided trajectories, strict filtering is applied to the rewritten trajectories. The filtering criteria include:
\begin{itemize}
    \item trajectories with incorrect final answers;
    \item trajectories that still explicitly rely on experience, as indicated by keywords such as 'refer to experience';
    \item trajectories that are excessively long, exceeding 6K in length;
    \item noisy trajectories containing long segments of meaningless strings.
\end{itemize}

\end{document}